\documentclass[journal]{IEEEtran}
\usepackage{tikz}
\usepackage{amsmath}
\pdfoutput=1
\usepackage{filecontents}
\usepackage{hyperref}       
\usepackage{url}            
\usepackage{booktabs}       
\usepackage{amsfonts}       
\usepackage{nicefrac}       
\usepackage{microtype}      
\usepackage{xcolor}         
\usepackage{multirow}
\usepackage{subcaption}
\usepackage{amsthm}
\usepackage{amsmath}
\usepackage{algorithm}
\usepackage{algorithmic}
\usepackage{enumerate}
\usepackage{caption}
\usepackage{graphicx}
\usepackage{float} 
\usepackage{booktabs}
\usepackage{verbatim}

\begin{document}
\newtheorem{theorem}{Theorem}
	\newtheorem{definition}{Definition}
	\newtheorem*{pf}{Proof}
        \newtheorem{lemma}{Lemma}
\title{\Large \bf F$^2$AT: Feature-Focusing Adversarial Training \\via Disentanglement of Natural and Perturbed Patterns}

 \author{Yaguan Qian, Chenyu Zhao, Zhaoquan Gu, Bin Wang, Shouling Ji, Wei Wang, Boyang Zhou, Pan Zhou



 }

\markboth{Journal of \LaTeX\ Class Files,~Vol.~14, No.~8, August~2021}%
{Shell \MakeLowercase{\textit{et al.}}: A Sample Article Using IEEEtran.cls for IEEE Journals}


\maketitle

\begin{abstract}
   Deep neural networks (DNNs) are vulnerable to adversarial examples crafted by well-designed perturbations. This could lead to disastrous results on critical applications such as self-driving cars, surveillance security, and medical diagnosis. At present, adversarial training is one of the most effective defenses against adversarial examples. However, traditional adversarial training is still difficult to achieve a good trade-off between clean accuracy and robustness since spurious features are still learned by DNNs. The intrinsic reason is that traditional adversarial training makes it difficult to fully learn core features from adversarial examples when adversarial noise and clean examples cannot be disentangled. In this paper, we disentangle the adversarial examples into natural and perturbed patterns by bit-plane slicing. We assume the higher bit-planes represent natural patterns and the lower bit-planes represent perturbed patterns, respectively. We propose a Feature-Focusing Adversarial Training (F$^2$AT), which differs from previous work in that it enforces the model to focus on the core features from natural patterns and reduce the impact of spurious features from perturbed patterns. The experimental results demonstrated that F$^2$AT outperforms state-of-the-art methods in clean accuracy and adversarial robustness.
\end{abstract}

\begin{IEEEkeywords}
Adversarial example, Adversarial training, Robustness, Trade-off.
\end{IEEEkeywords}

\section{Introduction}

\IEEEPARstart{D}espite the great success of deep neural networks (DNNs) in many tasks, such as object recognition  \cite{simonyan2014very, szegedy2015going}, machine translation \cite{chen2018best, sutskever2014sequence}, and speech translation \cite{xiong2016achieving, DBLP:journals/tist/ZhangGPMJS18}, they are susceptible to adversarial examples added by carefully designed imperceptible noises. This could lead to disastrous results on critical applications like self-driving cars, surveillance security, and medical diagnosis. To address this challenge, various defense methods have been proposed, such as feature compression methods \cite{DBLP:journals/corr/DziugaiteGR16, DBLP:journals/corr/DasSCHCKC17}, defensive distillation \cite{DBLP:conf/sp/PapernotM0JS16}, pruning \cite{DBLP:conf/iclr/DhillonALBKKA18}, and adversarial training \cite{DBLP:journals/corr/GoodfellowSS14, DBLP:conf/iclr/MadryMSTV18, DBLP:conf/iclr/KurakinGB17, DBLP:conf/iclr/TramerKPGBM18}.

Adversarial training has been proved as one of the strongest empirical defenses against adversarial attacks \cite{DBLP:journals/corr/abs-2102-01356}. The work in \cite{DBLP:journals/corr/abs-1810-06583} showed that the models obtained from adversarial training have significantly higher feature concentration, which tends to reduce the weight of non-predictive and weakly predictive features. Unfortunately, the model after recent adversarial training may still learn spurious features \cite{DBLP:conf/iclr/GeirhosRMBWB19, DBLP:conf/iclr/XiaoEIM21, DBLP:journals/corr/abs-2210-11369, NEURIPS2022_fd62b656}, which prevents further improvements to the model's performance. One possible reason is that previous adversarial training was directly conducted on adversarial examples, but did not notice the adversarial examples' internal composition. In this paper, we hope to learn more core features in adversarial training to further improve both clean accuracy and robustness.

        \begin{figure*}[t]
		\centering
			\includegraphics[width=0.75\linewidth]{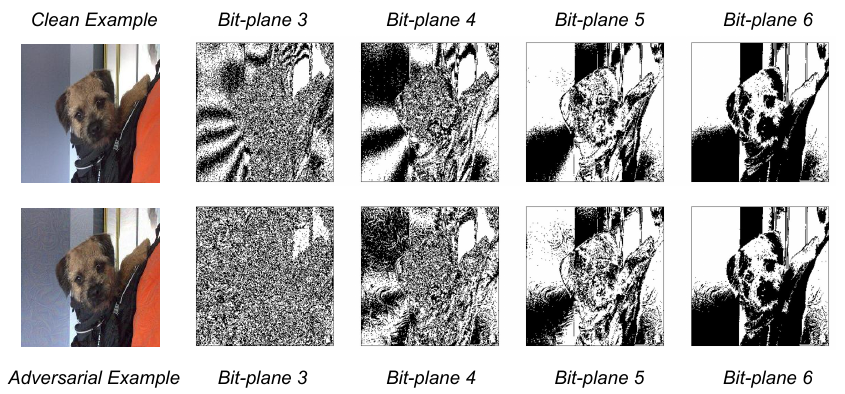}
		
			\caption{Bit-plane images of the clean and adversarial images. The information contained in the low bit-plane is more cluttered and contains more textural information. In contrast, the information contained in the high bit-plane is clearer and less susceptible to interference from adversarial perturbations. The black image represents the bit plane with all pixel values filled with zeros}
			\label{bp}
		
	\end{figure*}

Tabacof et al. \cite{DBLP:conf/ijcnn/TabacofV16} pointed out that adversarial examples are not necessarily isolated, spurious points, and that many adversarial images are located in dense regions of pixel space. Therefore, during adversarial training, we should perform some operations on the adversarial examples to reduce the effect of the adversarial example points in dense regions on the model learning the core features. Compared to clean examples, we found some extraordinary texture features occurring on the lower bit-planes of the adversarial example, but this phenomenon is not obvious on the upper bit-planes (see Fig. \ref{bp}). This finding inspires us to disentangle an adversarial image into two patterns, natural and perturbed patterns, by bit-plane slicing. Conceptually, an image's \textit{natural pattern} contains core features and its \textit{perturbed pattern} contains spurious features. In practice, the perturbed contains more detailed features like textures, which can be learned by models to improve the clean accuracy \cite{ilyas2019adversarial}. Specifically, upper bit-planes correspond to natural patterns, and lower bit-planes to perturbed patterns. We hope the model learns more core features in natural patterns and less spurious features in perturbed patterns to improve both the model's clean accuracy and adversarial robustness. 

To this end, we proposed a novel adversarial training method, named Feature-concentrate Adversarial Training (F$^2$AT), based on natural and perturbed patterns, which is different from existing adversarial training \cite{DBLP:journals/corr/GoodfellowSS14, DBLP:conf/iclr/MadryMSTV18, DBLP:conf/icml/ZhangYJXGJ19, DBLP:conf/iclr/0001ZY0MG20, DBLP:conf/nips/WuX020, DBLP:conf/cvpr/LeeLY20} in that they did not further disentangle adversarial examples to reduce the interference from adversarial noise. Through this disentanglement, on the one hand, we enforce the model to focus more on the natural patterns and less on perturbed patterns; on the other hand, we want the decision boundaries to be adjusted depending more on natural patterns while less on adversarial examples, which can avoid over-adjustment of the decision boundaries \cite{balaji2019instance}. Correspondingly, we designed two loss regularizers, \textit{i.e.}, pattern-dependent loss and natural-margin loss. (1) For \textit{pattern-dependent loss}, we employ mutual information (MI) to measure the dependence between the extracted features and the natural (perturbed) pattern, and maximize MI to expand the difference between the dependencies of the two patterns and the output. (2) For \textit{natural-margin loss}, the decision boundary constraint is used to expand the natural pattern to the decision boundary distance since natural patterns indicate more accurate position in pixel space than adversarial examples.
        
        F$^2$AT can improve both clean accuracy and adversarial robustness by learning more label-dependent strong predictive features. The experiments prove that the method is effective in defending against most adversarial attacks.

  	\textbf{Our contributions:}
  	
  	\begin{itemize}
  	\item We propose a novel disentanglement way for adversarial examples by bit-plane slicing. In this way, the adversarial example can be disentangled into natural and perturbed patterns, which provides the possibility for adversarial training to focus on core features to improve clean accuracy and robustness.
   
  	\item We propose a new adversarial training method based on this disentanglement. Two loss regularizers are developed to make the model focus on the core features of natural patterns. The benefits of this novel adversarial training can achieve a better trade-off between the clean accuracy and robustness of the target model.
   
  	\item The extensive experiments under both white-box and black-box settings are conducted on CIFAR-10 and CIFAR-100, which shows that clean accuracy and adversarial robustness with our adversarial training can be significantly improved than state-of-the-art methods.
    \end{itemize}

\section{Related Works}
	\textbf{Adversarial attacks.}
Since Szegedy et al. \cite{DBLP:journals/corr/SzegedyZSBEGF13} first proposed that adversarial examples can mislead deep neural networks, lots of effective adversarial attack methods, such as Fast Gradient Sign Method (FGSM) \cite{DBLP:journals/corr/GoodfellowSS14}, Projected Gradient Descent Attack (PGD) \cite{DBLP:conf/iclr/MadryMSTV18}, Carilini and Wagner Attack (CW) \cite{DBLP:conf/sp/Carlini017}, and Jacobian-based Saliency Map Attack (JSMA) \cite{JSMA} are proposed. Existing attack methods can be divided into white-box attacks and black-box attacks. White-box attacks are to know all the parameter information of the target model when generating adversarial examples, and black-box attacks are to know only part of the attacked model’s output when generating adversarial examples. In general, black-box attacks simulate the model gradient by repeatedly querying the target model (query-based attack) \cite{DBLP:conf/eccv/AndriushchenkoC20, DBLP:conf/eccv/BhagojiHLS18, DBLP:conf/ccs/ChenZSYH17, DBLP:journals/ijcv/WeiYL22, DBLP:conf/mm/YanW21} or searching for an alternative model similar to the target model (transfer-based attack) \cite{DBLP:conf/uss/DemontisMPJBONR19, DBLP:conf/iccv/HuangKGHBL19, DBLP:conf/iclr/LiuCLS17}. Since attackers hardly know the model parameters of the target model in practical applications, the model’s performance against black-box attacks can better reflect the real robustness. In \cite{DBLP:conf/ccs/Carlini017} and \cite{DBLP:conf/icml/AthalyeC018}, we have seen multiple defenses compromised by subsequent, more powerful attacks. This has motivated the research community to evaluate defenses against adaptive attacks \cite{DBLP:conf/nips/TramerCBM20}, which are specifically designed to trick defense mechanisms.

	\textbf{Adversarial defense.} Adversarial defense is divided into passive and active defense. Passive defense refers to the modification of model inputs to defend against adversarial attacks, e.g., denoising \cite{HGD} and detection of adversarial examples \cite{MagNet}. Active defense refers to the modification of model parameters to defend against adversarial attacks, \textit{e.g.}, adversarial training \cite{DBLP:conf/iclr/BaiZJXM021,features_not_bugs, DBLP:conf/iclr/Ma0WEWSSHB18, DBLP:conf/icml/ZhangYJXGJ19, DBLP:conf/icml/ZhangZ19}. Adversarial training has been proven to be one of the most effective defense methods \cite{DBLP:conf/iccv/ZiZMJ21}. In this paper, we focus on the defense of adversarial training.

Madry et al. \cite{DBLP:conf/eccv/BhagojiHLS18} formulate adversarial training as a minimax optimization problem. After that, much work has been proposed to further improve the robustness of adversarial training. For example, TRADES \cite{DBLP:conf/icml/ZhangYJXGJ19} achieves this by minimizing the dissimilarity between original examples and adversarial examples. MART \cite{DBLP:conf/iclr/0001ZY0MG20} enhances robustness by introducing a momentum term during training. 
	
	However, the adversarial examples in these adversarial training are typically generated from clean examples and share the same class labels. If both the labels of clean examples and adversarial examples are simultaneously utilized in the loss function as in the methods mentioned in literature \cite{DBLP:conf/icml/ZhangYJXGJ19} and \cite{DBLP:conf/iclr/0001ZY0MG20}, the model may learn to distinguish them based on the label information rather than genuinely capturing the underlying data characteristics \cite{DBLP:conf/iclr/KurakinGB17}. This can lead to an improvement in the model's performance on adversarial examples but may result in reduced robustness in real-world applications. Some methods can be applied to solve these problems. For example, SEAT \cite{DBLP:journals/corr/abs-2203-09678} does not invoke any regular terms, and robustness improvement can be achieved by smoothing the weights of historical models along the optimization trajectory. However, these methods make it difficult to avoid interference from spurious features of adversarial examples, making it difficult to learn more natural core features.
    
    Therefore, we propose a new adversarial training method that can make a model to focus on the core features of images by disentanglement of natural and perturbed patterns. The benefits of our method can improve the classification accuracy and robustness of the model at the same time.

\section{Preliminary}
In this paper, we use mutual information (MI) to measure the dependence between the extracted features and the natural (perturbed) pattern. MI can be thought of as the amount of information contained in one random variable about another random variable, or the reduction of uncertainty in one random variable due to another random variable. Before introducing the definition of mutual information, we first introduce the definition of entropy. Entropy is a measure of uncertainty of a random variable: $H(X) = \sum_{x}p(x)\log p(x)$ where $x$ is a specific value of $X$ and $p(x)$ is the marginal probability of $X$. The joint entropy of $X$ and $Y$, with joint mass probability $p(x, y)$, is the sum of the uncertainty contained by the two variables. Formally, joint entropy is defined as: $H(X,Y) = \sum_{x,y}p(x,y)\log p(x,y)$. The conditional entropy is defined as: 
\begin{align}\label{cond-entropy}
    H(X|Y) = - \sum_{x,y} p(x,y)\log p(x|y).
\end{align}

The definition of MI is based on the concept of entropy, comparing the difference between the entropy of individual variables and the joint entropy or conditional entropy. This quantifies the information transmission and association between variables. MI is defined as follows:
\begin{definition}
Let $X$ and $Y$ be random variables, the MI of ($X$, $Y$) is
    \begin{align}\label{d-1}
			I(X;Y)=\sum_{x,y}p(x,y)\log\frac{p(x,y)}{p(x)p(y)},
		\end{align}
\end{definition} 
 According to the definition, if $X$ and $Y$ are independent of each other, then $X$ provides no information about $Y$ and vice versa, $I(X; Y)=I(Y; X)=0$. At the other extreme, if $X$ is bijective to $Y$, then all the information passed is shared by $X$ and $Y$. In this case, $I(X; Y)=H(X)=H(Y)$. 

To better understand our proposed Theorem \ref{th2} in Section \ref{dmi}, we provide the following lemmas to explore the general relationship between MI and the information entropy:
\begin{lemma}
		Let $X$ and $Y$ be random variables, we have
            \begin{equation}
                \begin{aligned}
                    I(X;Y)&=H(X)-H(X|Y)\\
                    &=H(Y)-H(Y|X)\\
                    &=H(X)+H(Y)-H(X,Y),
                \end{aligned}\label{d1}
            \end{equation}
where $I(X;Y)$ denotes the MI of $X$ and $Y$, $H(X|Y)$ is the conditional entropy of $X$ and $Y$, $H(X,Y)$ is the joint entropy of $X$ and $Y$.
        \label{lem1}
	\end{lemma}

	\begin{lemma}
	Let $X$, $Y$, and $F$ be three random variables. We have:
	\begin{align}
		H(Y|X)=H(Y|X,F)+I(F;Y|X), \label{eq:d2-1}
	\end{align}
	\begin{align}
		I(X;F)=I(X;F|Y)+I(X;Y;F) \label{eq:d2-2}
	\end{align}
	\label{lem2}
    \end{lemma}
Information entropy and mutual information provide a precise mathematical tool to prove the rationality of our method. The strict proof of Lemmas \ref{lem1} and \ref{lem2} can be found in reference \cite{mceliece2002theory}

\section{Proposed Method}\label{method}

	
        \begin{figure*}
		\centering
            \includegraphics[width=0.9\textwidth]{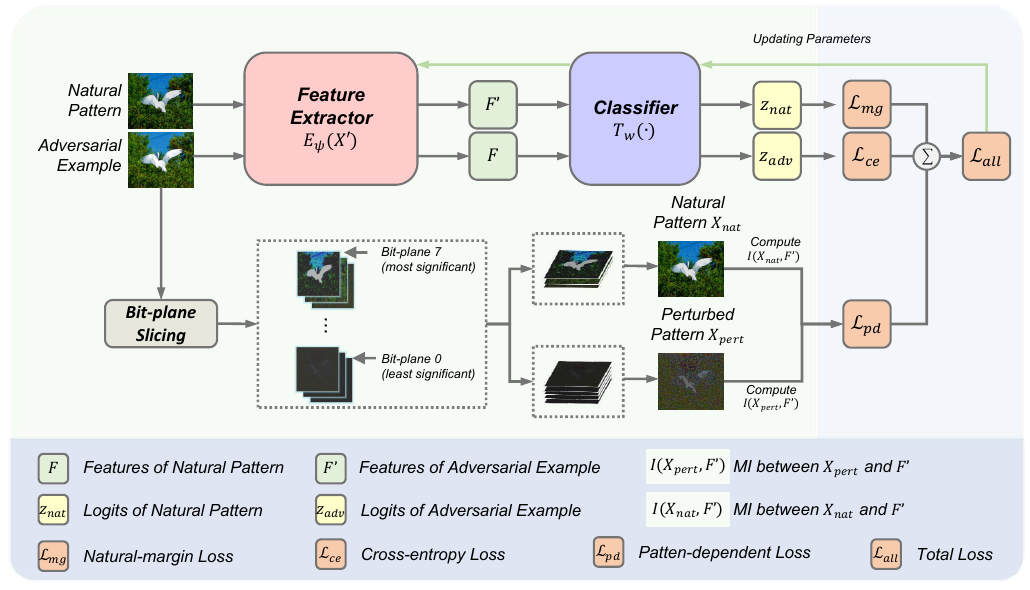}
		\caption{The overview of our proposed F$^2$AT. Compared to standard adversarial training (SAT) directly conducted on adversarial examples, we first disentangle adversarial examples into natural and perturbed patterns by bit-plane slicing. Through this disentanglement, we further conduct adversarial training on these two patterns. Accordingly, two loss functions, $\mathcal{L}_{pd}$ and $\mathcal{L}_{mg}$, are designed. For $\mathcal{L}_{pd}$, the model is forced to focus on the core features in the natural patterns. For $\mathcal{L}_{mg}$, we force the decision boundary to be far away from natural patterns.}

		\label{overview}
	\end{figure*}
\subsection{Motivation}\label{Mot}

Traditional adversarial training directly uses adversarial examples and does not disentangle them into perturbated and natural patterns, which will prevent adversarial training from further improving clean accuracy and robustness. In other words, the core features are masked by adversarial examples, making it difficult for adversarial training to focus on the core features. Bit-plane slicing provides an effective way to disentangle adversarial examples.

For an $n$-bit representation of a digital image, the spatial maps (of the same dimension as the image) corresponding to a specific bit position are known as an image's bit planes.  Bit plane $n$-1 corresponds to the most significant bit (MSB), and bit plane $0$ to the least significant bit (LSB).  The sum of $n$ bit planes weighted by their relative relevance can be thought of as an $n$-bit image.  In terms of pixel value and information content, characteristics included in lower bit planes are much less important than those embedded in higher bit planes \cite{shan2008image}. Lower bit planes are used in steganography techniques \cite{fridrich2001reliable} to incorporate critical copyright information that must be invisible to the human eye. However, information content in natural images decreases the MSB to the LSB. 

Large-scale features make up the majority of how people perceive images, and adding finer details only serves to amplify this view \cite{sugase1999global, sripati2009representing}. The human visual system naturally becomes resistant to the adversarial perturbations occurring at relatively lower bit planes because of this underlying knowledge of giving more weight to information present at the higher bit planes. Instead, these adversarial perturbations can arbitrarily change the categories in which deep network predictions fall, indicating a lack of baseline information that would otherwise make layering distinct bit planes in these networks important. As shown in Fig. \ref{bp}, the information contained in the low bit-plane is more cluttered and contains more textural information. On the contrary, the information contained in the high bit-plane is clearer and less susceptible to interference from adversarial perturbations.

Tabacof et al. \cite{DBLP:conf/ijcnn/TabacofV16} pointed out that adversarial examples are not necessarily isolated, spurious points, and that many adversarial images are located in dense regions of pixel space. Therefore, we should first perform some operations on the adversarial examples to reduce the effect of the adversarial example points in dense regions on the model learning the core features. We disentangle an image into two patterns, natural and perturbed patterns, by bit-plane slicing. That is to say, upper bit-planes are corresponding to natural patterns and lower bit-planes to perturbed patterns. We hope the DNN learns more about the \textit{core} features in natural patterns and less about the \textit{spurious} features in perturbed patterns to improve both the model's clean accuracy and adversarial robustness. 


\subsection{Natural and Perturbed Patterns}\label{NPP}

Let $\mathcal{X}$ be a natural image space, and $\mathcal{Y}$ be a label space, existing a ground-truth labeling function: $g: \mathcal{X}\rightarrow \mathcal{Y}$. The classification model $h$ is to approximate the ground-truth labeling function $g$. Let $h(X;w,\psi)=T_w\circ E_\psi(X)$, where $X\in \mathcal{X}$ is an image, $w$ and $\psi$ are the model parameters. The function $E_\psi:\mathcal{X} \rightarrow \mathcal{F}$ represents a feature extractor where $\mathcal{F}$ denotes the feature space. The function $T_w: \mathcal{F}\rightarrow \mathcal{Y}$ represents a classifier. Suppose ${X}'$ is an adversarial example generated on ${X}$, \textit{i.e}, for a clean image $X \in \mathcal{X}$, its adversarial example ${X}'=X+\delta$, such that $h(X')\neq y$ but $g(X')=y$, where $y \in \mathcal{Y}$ is the ground-truth label of $X$, and $\delta$ is the adversarial perturbation with constraint ${\Vert \delta \Vert}_p < \epsilon$. Here, ${\Vert \delta \Vert}_p$ is the $p$-norm and $\epsilon$ is a perturbation budget. Next, we formally define natural and perturbed patterns as follows:

\begin{definition}
Natural patterns. For an adversarial image $X'$, its natural pattern $X_{nat}$ refers to the component of the pixels employed by a feature extractor $E_\psi$. The natural pattern not only correlates with the ground-truth label $y$ but also are relevant to a ground-truth labeling function $g$.

\end{definition}
\begin{definition}
Perturbed patterns.  The contaminated part of the pixels is referred to as a perturbed pattern of $X'$, denoted by $X_{pert}$, which will interfere with a feature extractor $E_\psi$ and mislead a classifier to incorrect class labels.
\end{definition}

According to the above definitions, it is difficult to strictly disentangle an adversarial example into natural and perturbed patterns. Recall that lower bit-planes are easily contaminated by adversarial perturbation in Section~\ref{Mot}. Bit-plane slicing provides a feasible way to disentangle adversarial examples into natural and perturbed patterns.
Suppose $X$ is an $R$-bit image, we set the hyperparameter $K$ to slice the image. We accumulate the 0$\sim$($R-K-1$) bit-planes of the three RGB channels to get the \textit{perturbed pattern} $X_{pert}$ and add up the remaining upper bit-planes to obtain the \textit{natural pattern} $X_{nat}$. Note that for $X_{pert}$, $K\sim N$ bit-planes are filled with zero. Similarily, for $X_{nat}$, 0$\sim$($R-K-1$) bit-planes are filled with zero. Accordingly, we have $X'=X_{nat} + X_{pert}$.

\subsection{Feature Dependence}\label{dmi}
     
We expect the adversarial training to make the model focus more on the core features of natural patterns and less on the spurious features of perturbed patterns. In other words, we hope to reduce the noise interference from adversarial examples. In the context of DNNs, features can be referred to as low-level features like pixels, or high-level semantic features. For clarity, the features in this paper refer to the features obtained by a feature extractor for classifier: 

\begin{definition}

Classification feature. Let $E_\psi$ be a feature extractor composed of convolution layers in the context of DNNs. For an input image $X$, the output of the feature extractor is $F = E_\psi(X)$, $F\in \mathcal{F}$. Here $F$ is named classification features, which will be directly employed by a classifier $T_w$. When the context is clear, we refer to classification features simply as features.

\end{definition}
     
Considering mutual information (MI) is a good measure of the dependence between two sets \cite{DBLP:conf/icml/ZhouW00WZL22}, we intend to adopt MI to explicitly measure the dependence between an adversarial image $X'$ and its feature $F'$. Because our goal is to reduce the dependence on perturbed patterns in the adversarial training, we use $-I(X'; F')$ as a regular term added to the loss function. However, we find simply minimizing $-I(X'; F')$ in the loss function cannot achieve our expected goal. Since the loss does not guide the model to make a distinction between the learning of core features and noise. For this reason, we disentangle an adversarial image into perturbed and natural patterns by bit-plane slicing, \textit{i.e.}, $X'=X_{nat}+X_{pert}$, as described in Section \ref{NPP}. Similarly, we disentangle $I(X'; F') $ into $ I(X_{nat}; F') $ and $I(X_{pert}; F')$ for better guidance, where $ I(X_{nat}; F') $ and $I(X_{pert}; F')$ represent the MI of the natural and perturbed patterns to the adversarial example feature maps, respectively. We will prove the feasibility of this disentanglement by the bit-plane slicing to approximate MI linearly. Zhou et al. \cite{DBLP:conf/icml/ZhouW00WZL22} has given the transformation relation of MI in four variables as shown in Theorem \ref{t1}. 
    
\begin{theorem}\label{t1}
		Let $X'$, $X_{nat}$, $X_{pert}$, $F'$ be four random variables, such that $X' = X_{nat} + X_{pert} $, we have
    \begin{small}
        \begin{align}
        I(X';F')&=I(X_{nat};F')+I(X_{pert};F')-H(F'|X') \nonumber \\ 
        & \quad +H(F'|X_{nat},X_{pert})-I(X_{nat};X_{pert};F').
	\end{align}
    \end{small}	
	\end{theorem}
    
Unfortunately, computing $H(F'|X')$, $H(F'|X_{nat},X_{pert})$, and $I(X_{nat};X_{pert};F')$ is non-trivial \cite{DBLP:journals/neco/Paninski03}. Therefore, we tend to construct a surrogate for $I(X'; F')$. We find if $X' = X_{nat} + X_{pert}$ is implemented through bit-map slicing, $I(X'; F')$ can be lineally approximated by $I(X_{nat};F')$ and $I(X_{pert};F')$. Theorem \ref{th2} provides evidence to implement this idea as follows:

\begin{theorem} \label{th2}
Let ${X}'$ be an adversarial example, and $F'$ be the features of $X'$. If $X_{nat}$ and $X_{pert}$ are the natural and perturbed patterns of the adversarial examples generated by bit-plane slicing, respectively, then we have
		\begin{align}\label{t2}
			-H(F'|X')+H(F'|X_{nat}, X_{pert})-I(X_{nat}; X_{pert}; F') \approx 0.
		\end{align}
	
	\end{theorem}

\begin{pf}

        According to the definition of conditional entropy in Eq. \eqref{cond-entropy}, we have
        \begin{equation}\label{HFXX}
        \begin{aligned}
            &\quad H(F'|X_{nat}, X_{pert}) \\&= -\sum_{f',x_{nat},x_{pert}} p(f',x_{nat},x_{pert}) \log_2 p(f'|x_{nat},x_{pert}),
        \end{aligned}
        \end{equation}
where $p(f',x_{nat},x_{pert})$ represents the joint probability when $F'$, $X_{nat}$, and $X_{pert}$ take specific values $f'$, $x_{nat}$, and $x_{pert}$, and $p(f'|x_{nat},x_{pert})$ is the conditional probability of $F'$ given $x_{nat}$ and $x_{pert}$. 

Similarly, we have
\begin{equation} \label{HFX}
H(F'|X') = -\sum_{f',x'} p(f',x') \log_2 p(f'|x').
\end{equation}
Notice that for each adversarial example, there is a unique counterpart of the natural and perturbed pattern, i.e., $X' = X_{nat} + X_{pert}$ is established. We can express the probability of $X'$ in terms of $X_{nat}$ and $X_{pert}$, that is, $p(X') = p(X_{nat}, X_{pert})$. Substituting this relation into Eq.~\eqref{HFX}, we have
\begin{align}
H(F'|X') & = -\sum_{f',x'} p(f',x') \log_2 p(f|x') \nonumber \\
         & = -\sum_{f',x_{nat},x_{pert}} p(f',x_{nat},x_{pert}) \log_2 p(f'|x'). 
\end{align}
This is exactly the same as Eq.~\eqref{HFXX}. Therefore, we obtain
\begin{equation}
H(F|X_{nat}, X_{pert}) = H(F|X').
\end{equation}
It means $-H(F|X')+H(F|X_{nat}, X_{pert})=0$ in Eq.~\eqref{t2}. Next, we should prove $I(X_{nat}; X_{pert}; F') \approx 0$ in Eq.~\eqref{t2}. 

When $|I(X_{nat};X_{pert})-c|\leq \Delta$, where $c$ is a constant, $\Delta \ll c$, and $I(X_{nat};F')\neq 0$, ${I(X_{pert};F')\neq 0}$, we have:

For any given $\delta >$ 0, there exists a $\delta' >$ 0 such that for all $X_{pert}$ satisfying $I(X_{pert}, F')<\delta'$, it implies $I(X_{nat}, F')>\delta$, where $\delta'$ is a constant as small as possible.

Further applying the relationship between MI and entropy according to Lemma \ref{lem2} (Eq. \eqref{eq:d2-1} and \eqref{eq:d2-2}), we have
\begin{equation}
\begin{aligned}
&\quad I(X_{nat};X_{pert};F') \\&=I(X_{pert};F') - I(X_{pert};F'|X_{nat}) \\
&=I(X_{pert};F')-\left(H(X_{pert}|X_{nat})-H(X_{pert}|X_{nat}F')\right)\\
&\leq H(X_{nat}|X_{pert}F')+I(X_{pert};F')-H(X_{pert}|X_{nat})\\
&\leq H(X_{nat}|F')+I(X_{pert};F')-H(X_{pert}|X_{nat})\\
&=H(X_{nat})-I(X_{nat}|F')+I(X_{pert},F')-H(X_{pert}|X_{nat}).
\end{aligned}
\end{equation}
   
Taking $\delta=H(X_{nat})-H(X_{pert}|X_{nat})$, it holds that $I(X_{nat};X_{pert};F')\leq\delta'$. As $\delta'$ tends to 0, $I(X_{nat};X_{pert};F')$ also tends to 0.$\hfill\square$
\end{pf}



According to Theorem \ref{th2}, we can obtain an approximate of $I(X';F')$ by a linear combination of $I(X_{nat};F')$ and $I(X_{pert};F')$ as follows:
    \begin{align}\label{appr}
    I(X';F') \approx I(X_{nat};F')+I(X_{pert};F').
    \end{align}
When we increase $I(X_{nat}; F')$, $I(X'; F')$ increases accordingly due to the linearity of Eq.~\eqref{appr}. This allows the model to give different levels of attention to natural and perturbed patterns. We can simply set a trade-off hyperparameter to capture the degree of attention to the two patterns. 

\subsection{Feature-Focusing Adversarial Training}

Madry et al. \cite{DBLP:conf/eccv/BhagojiHLS18} formulate adversarial training as a minimax optimization problem formulated as follows:
\begin{align}\label{eq1}
	\min_{\theta}E_{(x,y)\sim D}[\max_{\delta\in\Omega}\mathcal{L}(h(x+\delta;\theta),y)],
\end{align}
where $h$ represents a deep neural network, $\theta$ represents the weight of $h$, $D$ represents a distribution of the clean example $x$ and the ground truth label $y$. $\mathcal{L}(h(x+\delta;\theta),y)$ represents loss function of updating the training model. $\delta $ represents the adversarial perturbation, and $\Omega $ represents a bound, which can be defined as $\Omega=\{\delta\colon\|\delta\|\leq\epsilon\}$ with the maximum perturbation strength.

	As mentioned in Section \ref{dmi}, the mutual information $I(X'; F')$ of an adversarial image $X'$ and its feature $F'$ is disentangled into a linear approximation of  $I(X_{nat}; F')$ and $I(X_{pert};F')$. This provides the possibility to force the model in adversarial training to pay different levels of attention to the natural and perturbed patterns when ensuring the dependence of the adversarial example on its features. Note that the perturbed mode is not completely useless and contains some detailed features that help improve clean accuracy. Recall that our goal is to guide the model to learn more core features, so we design two adversarial training strategies: (1) the feature concentration to make the model focus on the core features, and (2) the decision boundary separation to make the model focus on the classification features in natural patterns.
 
	\textbf{Pattern-dependent loss.} To learn more core features in natural patterns and reduce the impact of spurious features in perturbed patterns, 
    an intuitive way is to weigh the focus on these two patterns by optimizing the following loss function:
	\begin{align}\label{f2}
	   \mathcal{L} =-I \left(X_{nat}; E_\psi(X')\right)-\beta I \left(X_{pert}; E_\psi(X')\right),
	\end{align}
where $\beta$ is to weigh the importance of $I((X_{nat}; E_\psi(X'))$ and $I((X_{pert}; E_\psi(X'))$ on the prediction results. However, we found the estimate of $I((X_{nat}; E_\psi(X'))$ is much larger than that of $I((X_{pert}; E_\psi(X'))$, and there is a positive correlation between them. This indicates that adjusting the weights of $I \left(X_{pert}; E_\psi(X')\right)$ alone will not bring a significant change in the overall loss, which does not achieve our desired goal. To address this issue, we need to further expand the difference in the dependence of the output on the natural and perturbed patterns. Thus, pattern-dependent loss $\mathcal{L}_{pd}$ is defined, the loss is as follows:
\begin{equation}
\begin{aligned}
    \mathcal{L}_{pd} &= \mathbb{E}_\mathbb{P}\biggl[-\frac{1}{\tau}I\bigl(X_{nat}; E_\psi(X')\bigr) \quad + \\
    &\quad\log\sum_{i=1}^N\exp\left(\frac{I\left(X_{pert}^{(i)}; E_\psi(X')\right)}{\tau}\right)\biggr],
\end{aligned}
\end{equation}
where $\mathbb{P}$ is the probability distribution of $X$ and $\mathbb{E_\mathbb{P}}(\cdot)$ denotes the mathematical expectation, $N$ represents the number of randomly selected adversarial examples. Here, $\tau$ is a temperature coefficient. The larger $\tau$ is, the smoother the distribution of MI corresponding to the natural and perturbed patterns, and the loss will treat the natural and perturbed patterns equally. The smaller $\tau$ is, the more the model will focus on the natural patterns, but when $\tau$ is too small, the model will negate all dependencies with perturbation patterns.
\begin{figure}[htbp]
	\centering
	\begin{subfigure}{0.45\linewidth}
		\centering
		\includegraphics[width=\linewidth]{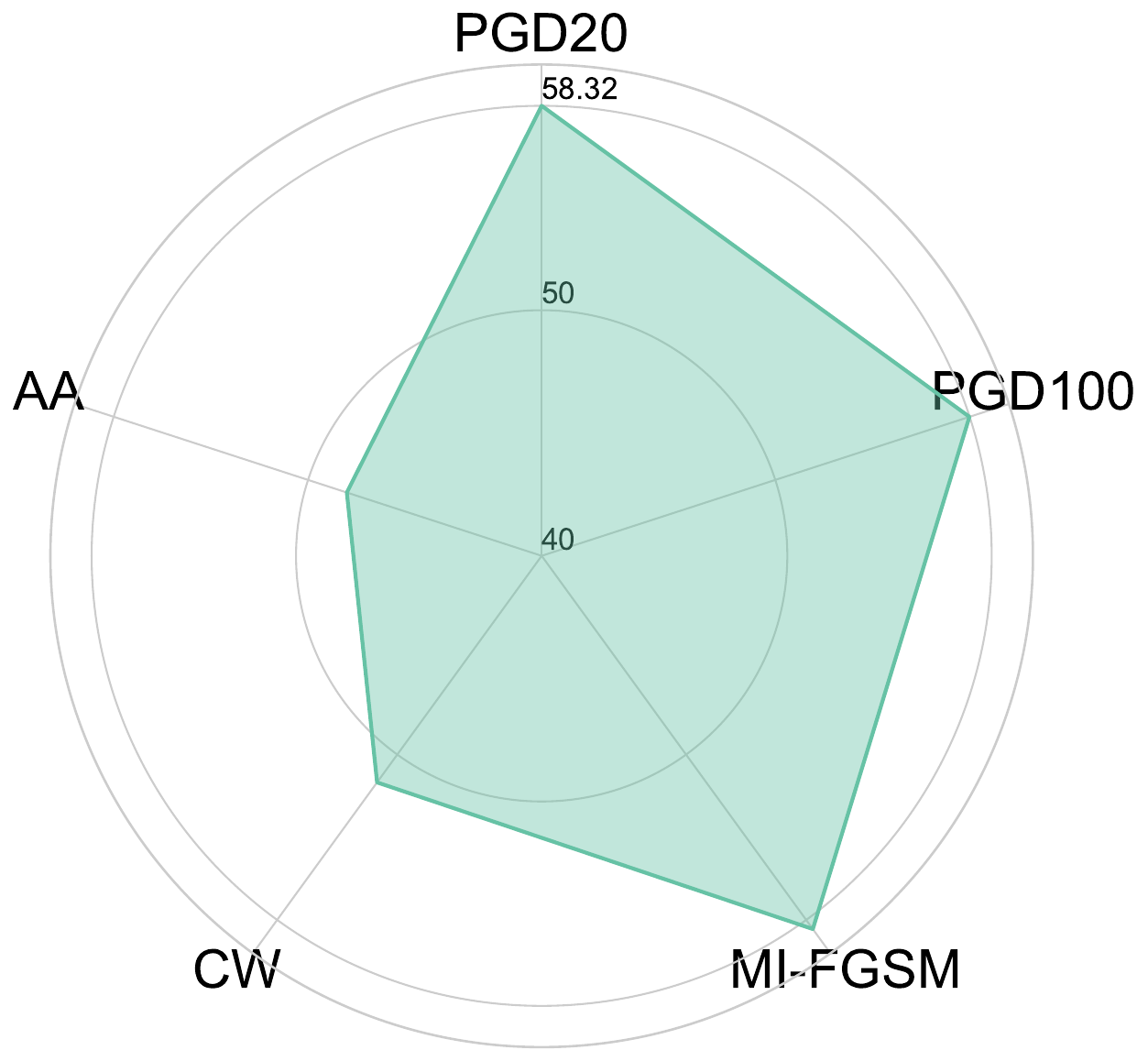}
		\caption{$\mathcal{L}_{ce}+\mathcal{L}_{mg}$}
		\label{hard_radar}
	\end{subfigure}
	\centering
	\begin{subfigure}{0.45\linewidth}
		\centering
		\includegraphics[width=\linewidth]{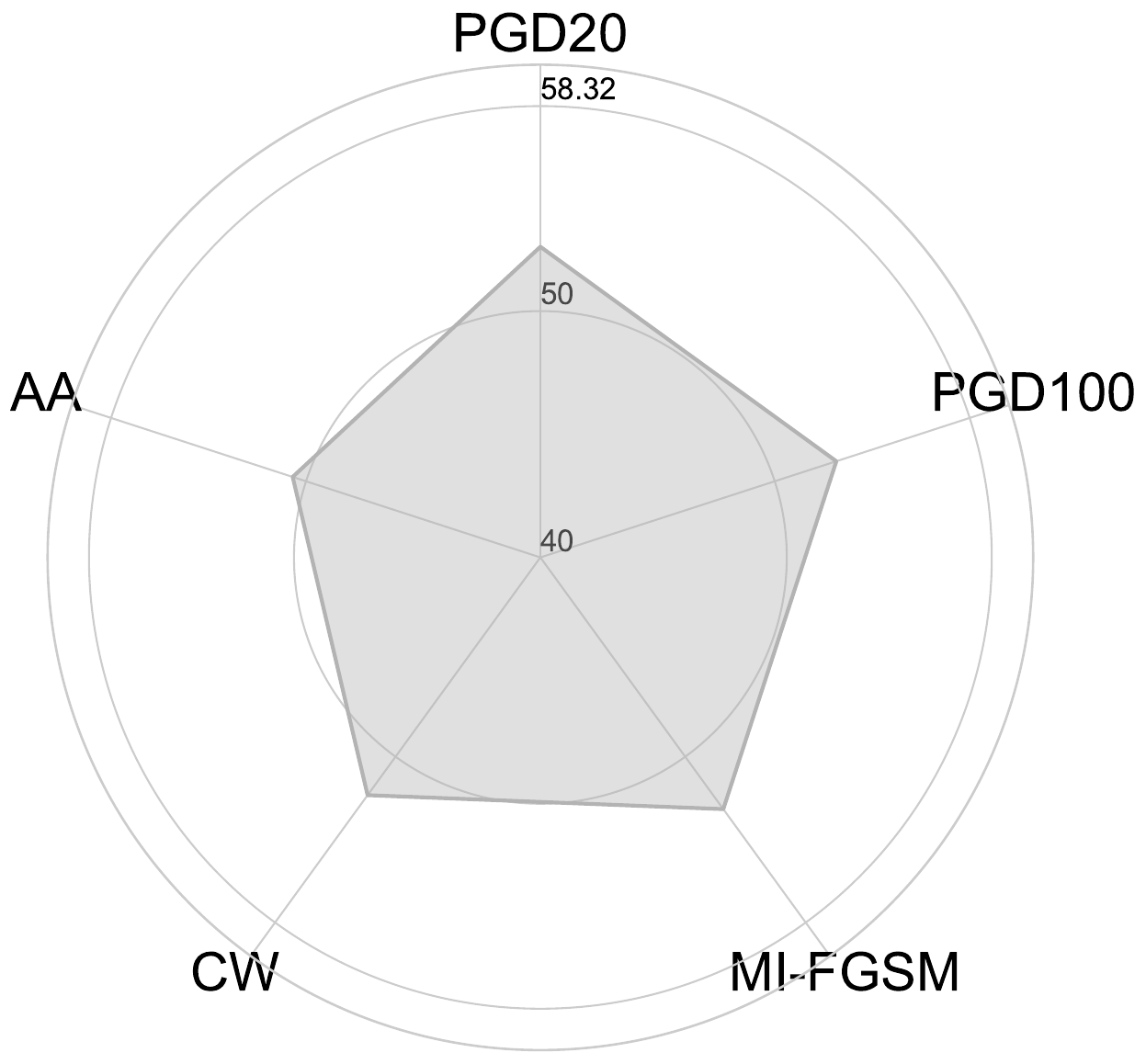}
		\caption{$\mathcal{L}_{ce}+\mathcal{L}_{mg}^{soft}$}
		\label{soft_radar}
	\end{subfigure}
        \caption{Adversarial accuracy of models trained against white-box attacks on CIFAR-10 using different losses. The target model is WideResNet-34-10. The models trained using $\mathcal{L}_{ce}+\mathcal{L}_{mg}$ are more accurate against weak attacks such as PGD and MI-FGSM, and less accurate against strong attacks such as CW and AA. The model trained using $\mathcal{L}_{ce}+\mathcal{L}_{mg}^{soft}$ is effective against both weak and strong attacks.}
        \label{radar}
        \end{figure}

\begin{algorithm}[!t]
		\caption{ Feature-Focusing Adversarial Training. }
		\label{alg:Framwork}
		\begin{algorithmic}[1] 
			\REQUIRE ~~\\ 
			Dataset $D$, training epochs $T$, batch size $N$, temperature coefficient $\tau$, hyperparameters $\alpha$, $\gamma$, $\upsilon$;
			\ENSURE ~~\\ 
            Model parameters $ w, \psi$.
			\STATE \textbf{for} $t = 1$ to $T$ \textbf{do}
			\STATE \quad \textbf{for} random batch $\{X^{(i)},y^{(i)}\}_{i=1}^N \sim D $ \textbf{do}
			\STATE \qquad Generate adversarial instance $X'^{(i)}$at the given perturbation budget $\epsilon$ for $X^{(i)}$ ;
			\STATE \qquad Slice the adversarial example $X'^{(i)}$ into $ X^{(i)}_{nat} $ and $ X^{(i)}_{pert} $ ;
			\STATE \qquad Calculate $\mathcal{L}_{all}$  using Eq. (\ref{Loss_all}) and optimize $ w, \psi $ by back-propagation;
			\STATE \quad \textbf{end for}
			\STATE \textbf{end for }
		
		\end{algorithmic}
	\end{algorithm}
        
\textbf{Natural-margin loss.}
The previous works \cite{DBLP:journals/corr/abs-1811-00525, DBLP:journals/corr/abs-1902-11019, DBLP:conf/nips/FawziMF16} claimed that adversarial examples are a natural consequence of learning decision boundaries. Therefore,  decision boundaries play a critical role in clean accuracy and robustness. Since the adversarial examples contain perturbed noise, traditional adversarial training tuning the decision boundaries according to the adversarial examples can improve robustness, but also decrease the clean accuracy due to its over-adjustment \cite{balaji2019instance}. In fact, the natural pattern of adversarial examples determines the clean accuracy (generalization) of the model's decision boundaries. Therefore, we propose a natural-margin loss to make the natural patters of adversarial examples as far as the decision boundaries.

	Define margin to measure the distance between different class decision boundaries, and maximize margin so that the decision boundaries of different classes are distinguished as much as possible. We define the margin as follows:
	\begin{align}\label{margin}
		mg(X',y)=\left[h\left(X_{nat}\right)\right]_y-\max\limits_{y'\neq y}\left[h\left(X_{nat}\right)\right]_{y'},
	\end{align}
 where $y$ and $y'$ are label indexes, and $\left[h\left(X_{nat}\right)\right]_y$ means the logit value corresponding to the natural patterns $X_{nat}$ labeled $y$. 
	 The positive value of $mg(X,y)$ in Eq. \eqref{margin} means the distance of the examples correctly classified margin value. However, we find that calculating the mean by Eq. \eqref{margin} cancels out the correctly classified margin value (positive number) and the misclassified margin value (negative number) in a mini-batch, and to solve this problem, the margin loss will be defined as:
	\begin{align}\label{Loss_mg}
		\mathcal{L}_{mg}=-\mathbb{E_\mathbb{P}}\left[\left[h\left(X_{nat}\right)\right]_y-\max[ h\left(X_{nat}\right)]\right],
	\end{align}
	where $\mathbb{P}$ is the probability distribution to the natural pattern and $\max[\cdot]$ means the maximum element of the vector. However, from Fig. \ref{hard_radar}, we found that models trained using $\mathcal{L}_{ce}+\mathcal{L}_{mg}$ were more accurate against weak attacks such as PGD and MI-FGSM, and less accurate against strong attacks such as CW and AA. We speculate that this is due to the overfitting of images generated for weak attacks. Therefore, we need to regularize the marginal loss. We further improve using the soft marginal loss of the natural model, which is defined as follows:
        \begin{align}\label{Loss_mg_soft}
		\mathcal{L}_{mg}^{soft}=-\mathbb{E_\mathbb{P}}\left[\left[h\left(X_{nat}\right)\right]_y-\frac{1}{\upsilon}\log \sum\limits_{y'\neq y}[ \upsilon \cdot \exp \left( h\left(X_{nat}\right)]\right)\right],
	\end{align}

    From Fig. \ref{soft_radar}, the model trained using $\mathcal{L}_{ce}+\mathcal{L}_{mg}^{soft}$ is effective against both weak and strong attacks. Then, we get the total loss of the adversarial training:
	\begin{align}\label{Loss_all}
		\mathcal{L}_{all}=\mathcal{L}_{ce}+\alpha \mathcal{L}_{pd}+\gamma \mathcal{L}^{soft}_{mg},
	\end{align}
	where $\mathcal{L}_{ce}$ is a cross-entropy loss widely used in deep learning. $\alpha$ and $\gamma$ are the trade-off hyperparameters, which play a critical role in balancing the importance of attention to natural and perturbed patterns for a trade-off between clean and robust accuracy. The details of the training are presented in Algorithm \ref{alg:Framwork}.
	

	\section{Experiments}
 \subsection{Experimental settings}
	\textbf{Baseline.}
	Our implementation, built upon the PyTorch framework, underwent rigorous experimentation across various benchmark datasets, notably CIFAR-10 \cite{Krizhevsky2009LearningML} and CIFAR-100 \cite{Krizhevsky2009LearningML}. The chosen target models were Wide-ResNet (WRN-34-10) \cite{DBLP:conf/bmvc/ZagoruykoK16} and ResNet-18 \cite{he2016deep}. In the white-box setup, we selected multiple random seeds for our experiments. In the black-box setting, we have chosen the best-performing model from the white-box setup for testing. For the surrogate model in the black-box setup, we chose VGG-16\cite{simonyan2014very}. To assess the effectiveness of our approach, we compared its performance against several baseline methods, including Madry et al.'s PGD-based adversarial training (SAT) \cite{DBLP:conf/iclr/MadryMSTV18}, TRADES \cite{DBLP:conf/icml/ZhangYJXGJ19}, MART \cite{DBLP:conf/iclr/0001ZY0MG20}, and SEAT \cite{DBLP:journals/corr/abs-2203-09678}. 
	
        \begin{table*}[htbp]
		\centering
		\caption{Adversarial accuracy (\%) of defense methods against white-box attacks on CIFAR-10. The target model is ResNet-18.}
  \resizebox{0.9\textwidth}{!}{
		\begin{tabular}{l|c|ccccc} 
			\toprule
			Defenses & None & PGD20 & PGD100 & MI-FGSM & CW & AA \\
			\midrule
			SAT \cite{DBLP:conf/iclr/MadryMSTV18} & 77.59 $\pm$ 0.62 & 45.45 $\pm $0.35 & 45.42 $\pm$ 0.42 & 46.32 $\pm$ 0.34 & 43.90 $\pm$ 0.63 & 42.93 $\pm$ 0.41\\
			TRADES \cite{DBLP:conf/icml/ZhangYJXGJ19} & 80.52 $\pm$ 0.88 & 49.41 $\pm$ 0.44 & 49.40 $\pm$ 0.38 & 50.02 $\pm$ 0.09 & 47.17 $\pm$ 0.46 & 46.06 $\pm$ 0.94 \\
                MART \cite{DBLP:conf/iclr/0001ZY0MG20} & 79.97 $\pm$ 0.62 & 50.59 $\pm$ 0.01  & 50.58 $\pm$ 0.02 &  50.43 $\pm$ 0.01 & 47.39 $\pm$ 0.60 & 46.18 $\pm$ 0.18 \\

                SEAT \cite{wang2022self} & 79.91 $\pm$ 0.28 & 48.79 $\pm$ 0.02  & 48.80 $\pm$ 0.02 & 49.19 $\pm$ 0.57 & 47.42 $\pm$ 0.02 & 45.39 $\pm$ 0.01  \\
                
                F$^2$AT(ours) & \textbf{80.56 $\pm$ 0.24} & \textbf{50.65 $\pm$ 0.19} & \textbf{50.66 $\pm$ 0.18} & \textbf{50.05 $\pm$ 0.27} & \textbf{47.81 $\pm$ 0.01} & \textbf{46.54 $\pm$ 0.27} 
		\\
			\bottomrule
		\end{tabular}}
		\label{tb1-CIFAR-10-res}
	\end{table*}
	
	\begin{table*}[htbp]
		\centering
		\caption{Adversarial accuracy (\%) of defense methods against white-box attacks on CIFAR-10. The target model is WRN-34-10.}
            \resizebox{0.9\textwidth}{!}{
		\begin{tabular}{l|c|ccccc} 
			\toprule
			Defenses & None & PGD20 & PGD100 & MI-FGSM & CW & AA \\
			\midrule
			SAT \cite{DBLP:conf/iclr/MadryMSTV18} & 85.61 $\pm$ 0.04 & 46.67 $\pm $0.03 & 46.74 $\pm$ 0.04 & 48.05 $\pm$ 0.20 & 47.01 $\pm$ 0.15 & 44.56 $\pm$ 2.04\\
			TRADES \cite{DBLP:conf/icml/ZhangYJXGJ19} & 85.45 $\pm$ 0.12 & 51.89 $\pm$ 2.00 & 51.01 $\pm$ 2.08 & 53.41 $\pm$ 1.07 & 52.10 $\pm$ 0.14 & 50.16 $\pm$ 1.77 \\
                MART \cite{DBLP:conf/iclr/0001ZY0MG20} & 83.70 $\pm$ 0.24& 53.96 $\pm$ 0.09  & 53.98 $\pm$ 0.11 &  55.14 $\pm$ 0.14 & 51.31 $\pm$ 0.22 & 49.44 $\pm$ 0.03 \\

                SEAT \cite{wang2022self} & 84.35 $\pm$ 0.30 & 52.73 $\pm$ 0.12  & 52.71 $\pm$ 0.13 & 53.82 $\pm$ 0.10 & 51.96 $\pm$ 0.08 & 50.05 $\pm$ 0.05  \\
                
                F$^2$AT(ours) & \textbf{85.64 $\pm$ 0.13} & \textbf{54.32 $\pm$ 1.22 }& \textbf{54.31 $\pm$ 1.26} & \textbf{55.65 $\pm$ 1.16 }& \textbf{52.39 $\pm$ 0.73} & \textbf{50.53 $\pm$ 0.07} 
		\\
			\bottomrule
		\end{tabular}}
		\label{tb1-CIFAR-10}
	\end{table*}
\begin{table*}[t!]
		\centering
		\caption{Adversarial accuracy (\%) of defense methods against white-box attacks on CIFAR-100. The target model is WRN-34-10.}
  \resizebox{0.9\textwidth}{!}{
		\begin{tabular}{l|c|ccccc} 
			\toprule
			Defenses & None & PGD20 & PGD100 & MI-FGSM & CW & AA \\
			\midrule
    	 	
			SAT \cite{DBLP:conf/iclr/MadryMSTV18} & 53.64 $\pm$ 0.47 & 24.49 $\pm$ 0.13 & 24.48 $\pm$ 0.16 & 24.97 $\pm$ 0.02 & 23.00 $\pm$ 0.01  & 21.01 $\pm$ 0.01\\
   
			TRADES \cite{DBLP:conf/icml/ZhangYJXGJ19} & 58.54 $\pm$ 0.14 & 29.92 $\pm$ 1.25 & 29.93 $\pm$ 1.04 & 30.27 $\pm$ 0.99 & 27.21 $\pm$ 0.39 & 25.96 $\pm$ 0.01 \\
                MART \cite{DBLP:conf/iclr/0001ZY0MG20} & 57.99 $\pm$ 0.25 & 28.89 $\pm$ 0.18 & 28.87 $\pm$ 0.20 & 29.43 $\pm$ 0.13 & 27.10 $\pm$ 0.28 & 25.74 $\pm$ 0.25 \\

                SEAT \cite{wang2022self} & 57.53 $\pm$ 0.39 & 29.95 $\pm$ 0.04 & 29.96 $\pm$ 0.05 & 30.59 $\pm$ 0.04 & 28.52 $\pm$ 0.01 & 26.17 $\pm$ 0.01  \\
                
                F$^2$AT(ours) & \textbf{60.21 $\pm$ 0.14} & \textbf{30.29 $\pm$ 0.12 }& \textbf{30.27 $\pm$ 0.13} & \textbf{30.87 $\pm$ 0.21 }& \textbf{28.89 $\pm$ 0.03} & \textbf{26.91 $\pm$ 0.01} 
		\\
			\bottomrule
		\end{tabular}}
		\label{tb1-CIFAR-100}
	\end{table*}

        \begin{table*}[!t]
		\centering
		\caption{Adversarial accuracy (\%) of defense methods against white-box attacks on CIFAR-100. The target model is ResNet-18.}
  \resizebox{0.9\textwidth}{!}{
		\begin{tabular}{l|c|ccccc} 
			\toprule
			Defenses & None & PGD20 & PGD100 & MI-FGSM & CW & AA \\
			\midrule
			SAT \cite{DBLP:conf/iclr/MadryMSTV18} & 51.83 $\pm$ 0.47 & 22.55 $\pm$ 0.13  & 22.56 $\pm$ 0.16 & 23.21 $\pm$ 0.02 & 21.26 $\pm$ 0.01 & 19.32 $\pm$ 0.03 \\

			TRADES \cite{DBLP:conf/icml/ZhangYJXGJ19} & 53.46 $\pm$ 0.86 & 25.98 $\pm$ 1.42 & 26.06 $\pm$ 1.38 & 26.32 $\pm$ 1.41 & 21.90 $\pm$ 1.62 & 21.96 $\pm$ 0.49 \\

                MART \cite{DBLP:conf/iclr/0001ZY0MG20} & 51.35 $\pm$ 0.02 & 26.28 $\pm$ 0.03 & 26.26 $\pm$ 0.02 & 26.69 $\pm$ 0.01 & 23.13 $\pm$ 0.01 & 21.81 $\pm$ 0.02 \\
 	 	 	 	
                SEAT \cite{wang2022self} & 53.74 $\pm$ 0.88 & 26.25 $\pm$ 0.08 & 26.21 $\pm$ 0.04 & 26.77 $\pm$ 0.13 & 24.22 $\pm$ 0.27 & 22.49 $\pm$ 0.21  \\
                 				
                F$^2$AT(ours) & \textbf{54.19 $\pm$ 0.79} & \textbf{26.75 $\pm$ 0.14 }& \textbf{26.78 $\pm$ 0.04} & \textbf{27.18 $\pm$ 0.02 }& \textbf{25.14 $\pm$ 0.11} & \textbf{23.24 $\pm$ 0.13 }
		\\
			\bottomrule
		\end{tabular}}
		\label{tb1-CIFAR-100-res}
	\end{table*}

 \begin{table*}[!t]
		\centering
		\caption{Adversarial accuracy (\%) of defense methods against white-box attacks on Tiny-ImageNet. The target model is ResNet-18.}
  \resizebox{0.9\textwidth}{!}{
		\begin{tabular}{l|c|ccccc} 
			\toprule
			Defenses & None & PGD20 & PGD100 & MI-FGSM & CW & AA \\
			\midrule
			SAT \cite{DBLP:conf/iclr/MadryMSTV18} & 31.52 $\pm$ 1.03 & 12.78 $\pm$ 0.10 & 12.80 $\pm$ 0.11 & 12.93 $\pm$ 0.11 & 10.48 $\pm$ 0.14 & 9.37 $\pm$ 0.17  \\
 	  	
			TRADES \cite{DBLP:conf/icml/ZhangYJXGJ19} & 36.19 $\pm$ 0.66 & 11.89 $\pm$ 0.08 & 11.86 $\pm$ 0.08 & 11.96 $\pm$ 0.04 & 8.51 $\pm$ 0.06 & 8.26 $\pm$ 0.21 \\
                MART \cite{DBLP:conf/iclr/0001ZY0MG20} & 33.57 $\pm$ 0.36  & 13.93 $\pm$ 1.13 & 13.93 $\pm$ 1.14 & 14.07 $\pm$ 1.14 & 11.25 $\pm$ 0.92 & 10.39 $\pm$ 0.77 \\
                SEAT \cite{wang2022self} & 35.51 $\pm$ 0.04 & 15.14 $\pm$ 0.27 & 15.16 $\pm$ 0.25 & 15.35 $\pm$ 0.19 & 12.61 $\pm$ 0.09 & 11.37 $\pm$ 0.06  \\
                 				
                F$^2$AT(ours) & \textbf{40.54 $\pm$ 0.03} & \textbf{16.30 $\pm$ 1.12 }& \textbf{16.32 $\pm$ 1.15} & \textbf{16.51 $\pm$ 1.12 }& \textbf{14.33 $\pm$ 0.85} & \textbf{13.13 $\pm$ 0.79 }
		\\
			\bottomrule
		\end{tabular}}
		\label{tb1-tiny-res}
	\end{table*}
	\textbf{Attack setting.}
    To illustrate the efficacy of our adversarial defense algorithm, we evaluate adversarial accuracy through both white-box and black-box adversarial attacks. This involves measuring the model's accuracy under diverse adversarial attack scenarios and calculating the average robustness on the test set. For white-box attacks, we primarily employ four attack methods: PGD \cite{DBLP:conf/iclr/MadryMSTV18}, MI-FGSM \cite{DBLP:conf/cvpr/DongLPS0HL18}, CW \cite{DBLP:conf/sp/Carlini017}, and AA \cite{DBLP:conf/icml/Croce020a}. Meanwhile, black-box attacks consist of six common methods: FGSM \cite{DBLP:journals/corr/GoodfellowSS14}, PGD \cite{DBLP:conf/iclr/MadryMSTV18}, TI-FGSM \cite{dong2019evading}, MI-FGSM \cite{DBLP:conf/cvpr/DongLPS0HL18}, DI-FGSM \cite{xie2019improving}, and VMI-FGSM \cite{wang2021enhancing}. The perturbation budgets for white-box and black-box attacks are 8/255. Additionally, PGDT represents a PGD attack with T iteration steps and a step size of $0.007$.
    \begin{figure}[t]
		\centering
		\includegraphics[width=0.48\textwidth]{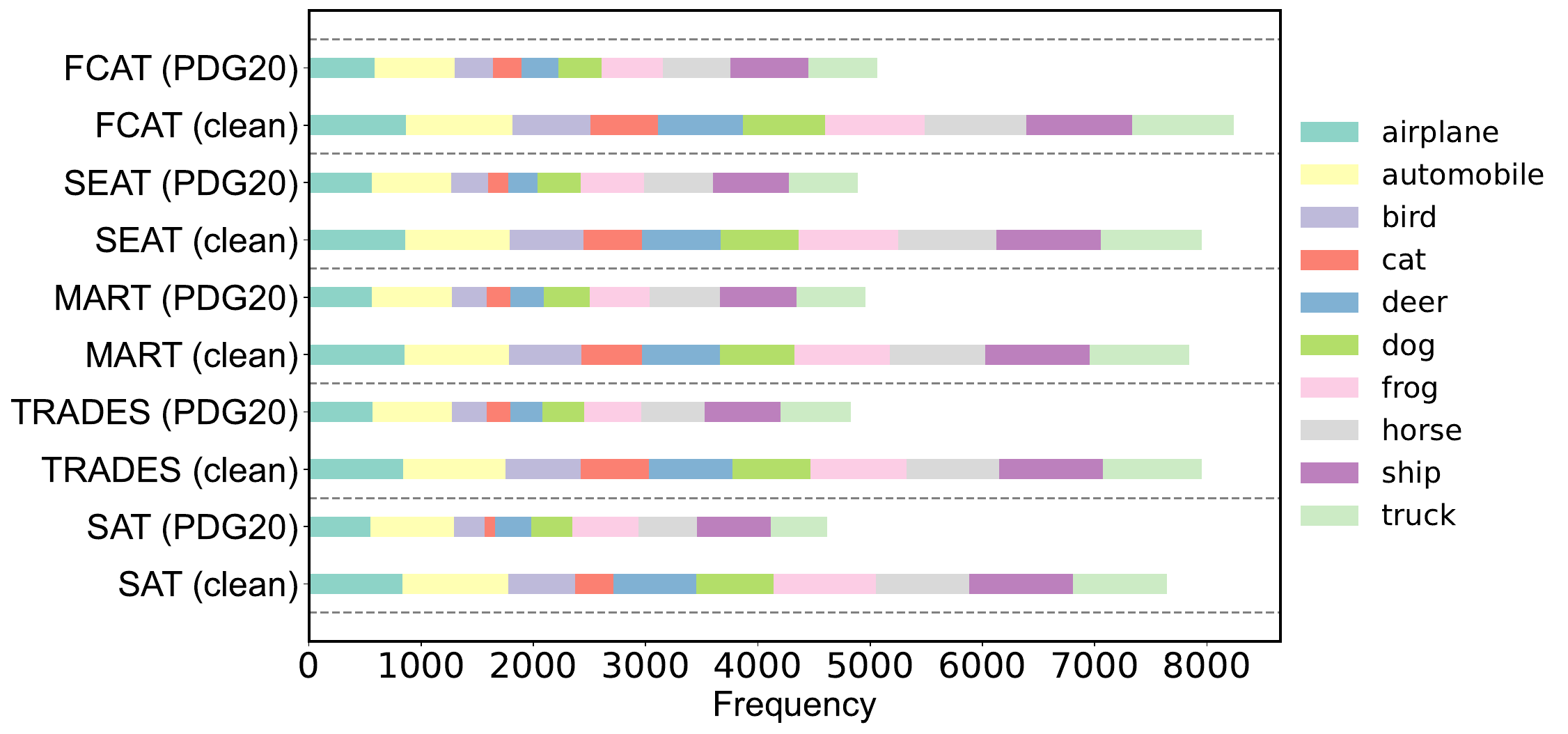}
		\caption{Distribution of the correct predictions of different training methods for each class on CIFAR- 10 using ResNet-18. We can infer that our method predicts the frequency of each class more evenly.}
		\label{res-Frequency}
	\end{figure}
 \begin{figure*}[!t]
        \centering
	\begin{subfigure}{0.24\linewidth}
		\centering
		\includegraphics[width=1.1\linewidth]{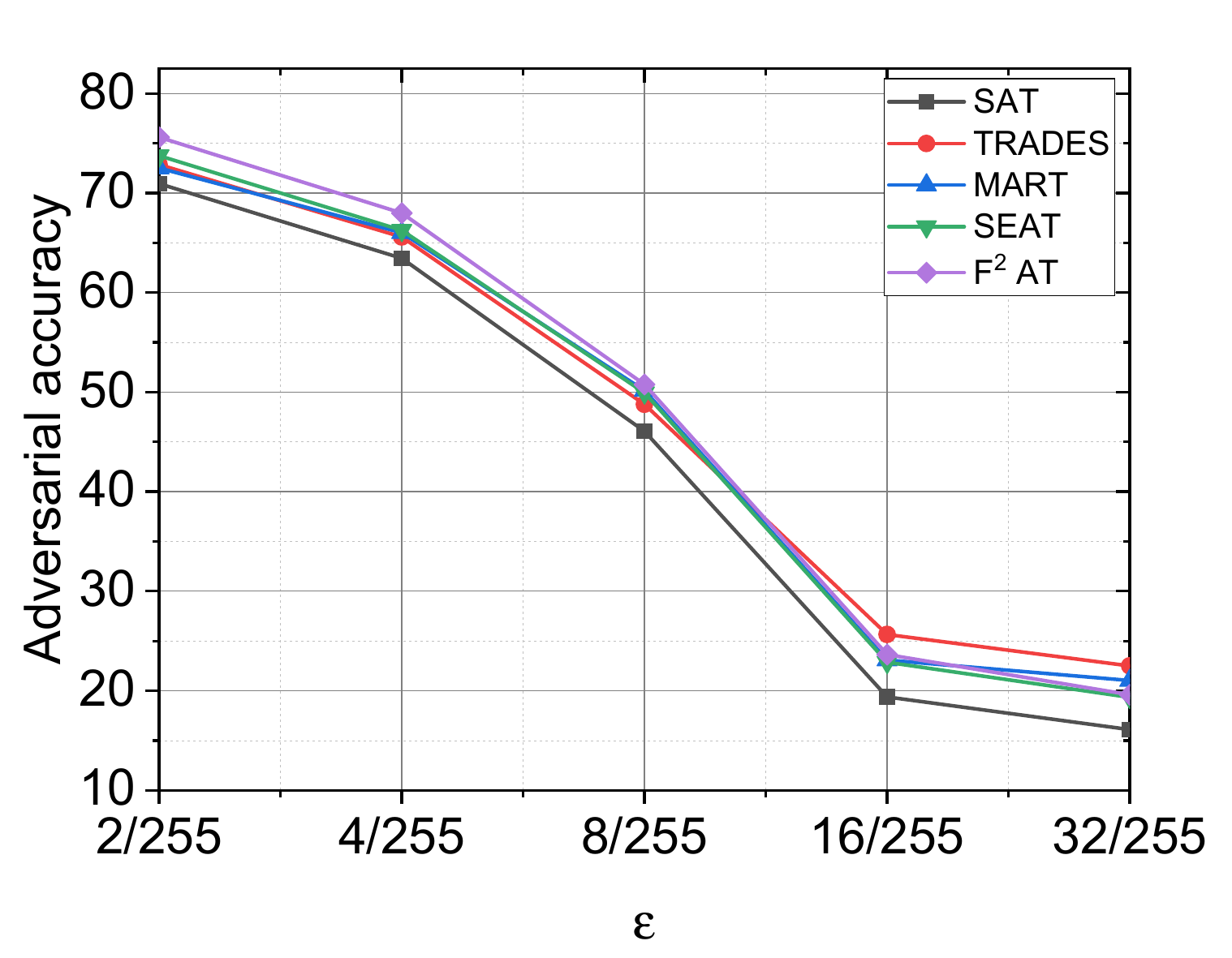}
		\caption{ResNet-18}
		\label{res-epi}
	\end{subfigure}
	\centering
	\begin{subfigure}{0.24\linewidth}
		\centering
		\includegraphics[width=1.1\linewidth]{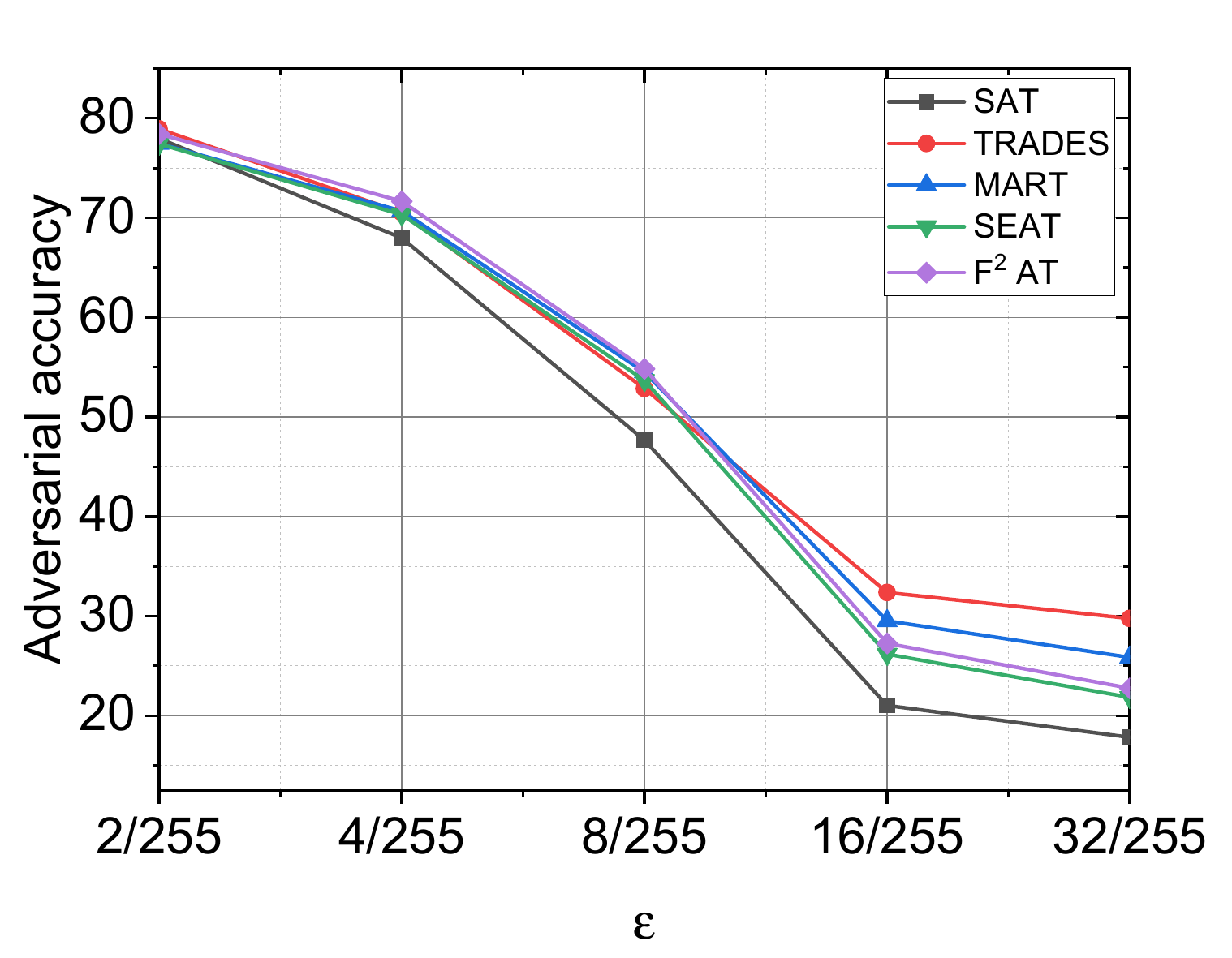}
		\caption{WRN-34-10}
		\label{wrn-epi}
	\end{subfigure}
	\centering
	\begin{subfigure}{0.24\linewidth}
		\centering
		\includegraphics[width=1.1\linewidth]{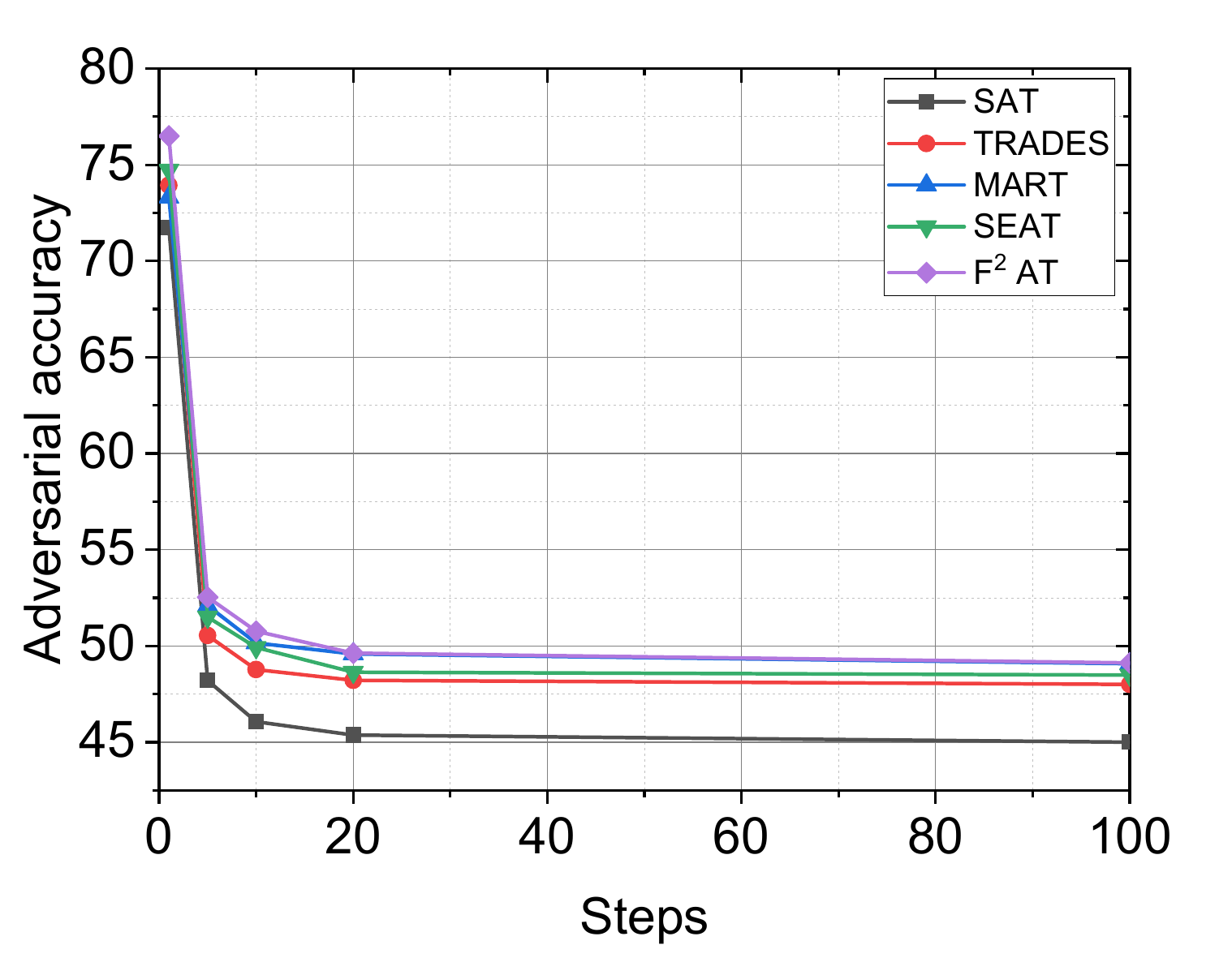}
		\caption{ResNet-18}
		\label{res-steps}
	\end{subfigure}
	\centering
	\begin{subfigure}{0.24\linewidth}
		\centering
		\includegraphics[width=1.1\linewidth]{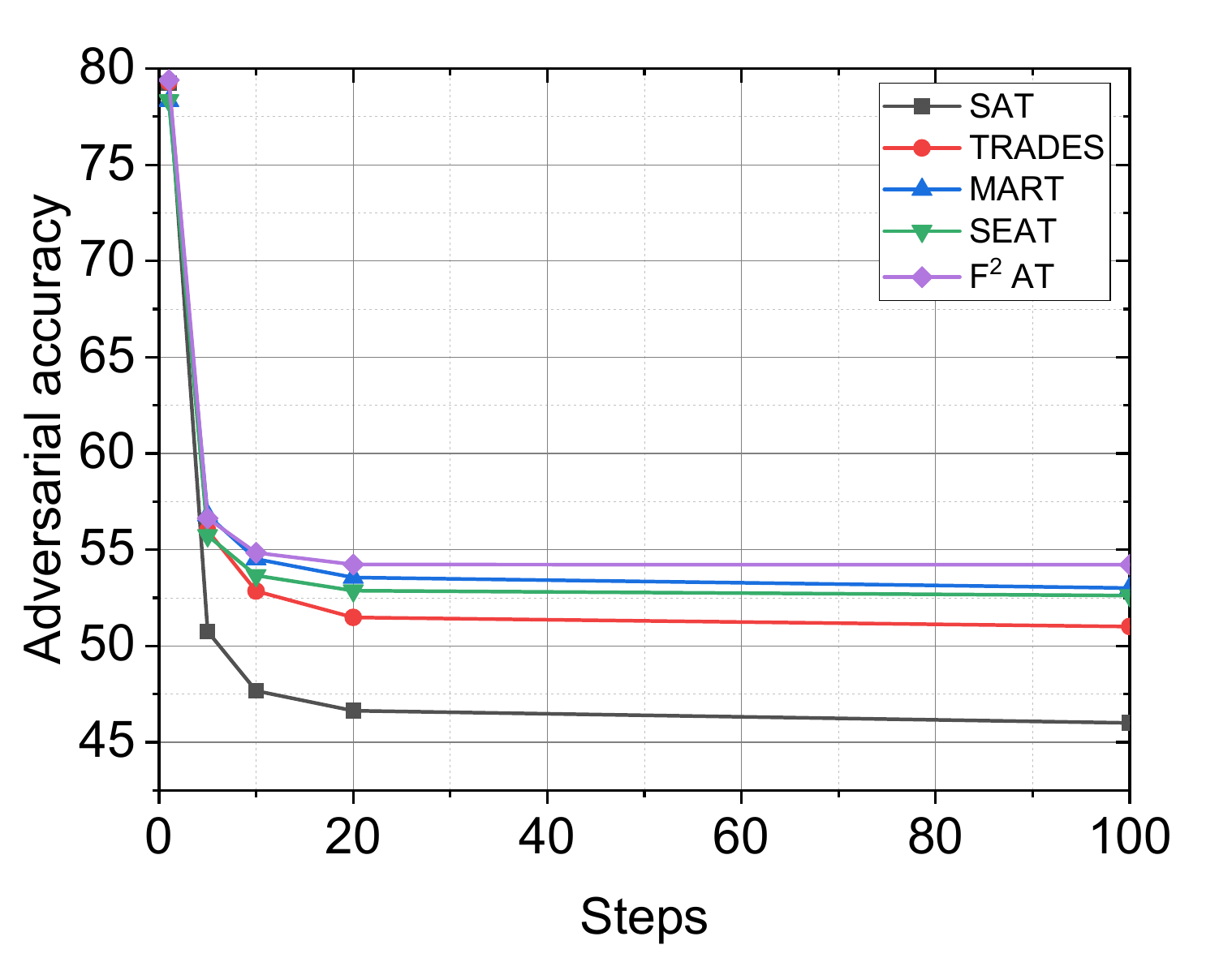}
		\caption{WRN-34-10}
		\label{wrn-steps}
	\end{subfigure}
        \caption{The effects of iteration steps and perturbation budgets $\epsilon$. The attack used is a PGD attack with an iterative step size of 0.007. The steps are set to 10 when exploring the effect of $\epsilon$, and the $\epsilon$ is set to 8/255 when exploring the effect of steps. (a) and (b) show that our method is able to achieve good robustness in the face of perturbations below 8/255. However, when the perturbation size increases to 16/255, the robustness decreases and is lower than that of TRADES and MART. As obtained from (c) and (d), our method outperforms the other methods in all settings of steps.}
        \label{epi-steps}
        \end{figure*}
        
         \begin{figure}[!t]
	\centering
	\begin{subfigure}{0.75\linewidth}
		\centering
		\includegraphics[width=0.9\linewidth]{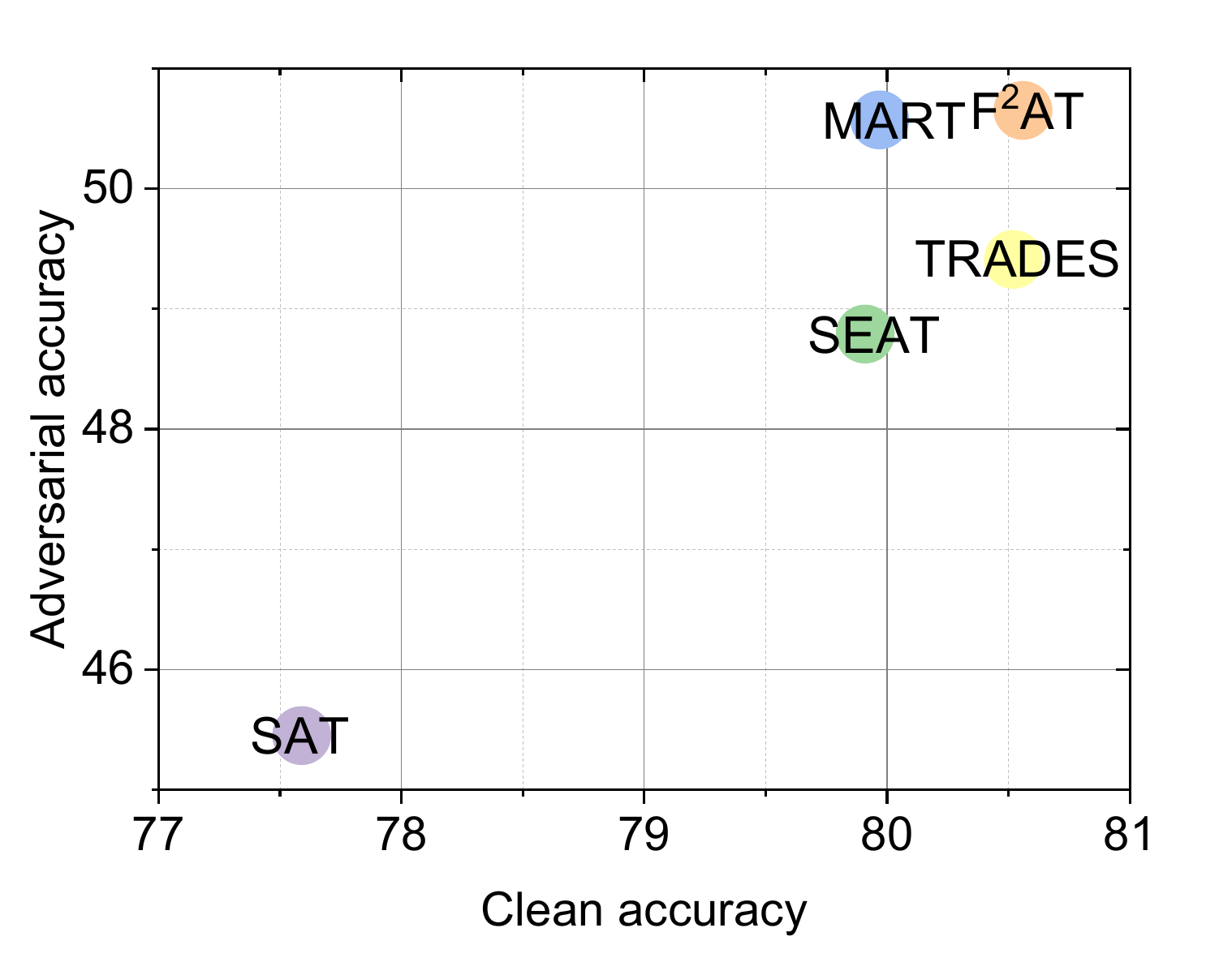}
		\caption{ResNet-18}
		\label{res}
	\end{subfigure}
	\centering
	\begin{subfigure}{0.75\linewidth}
		\centering
		\includegraphics[width=0.9\linewidth]{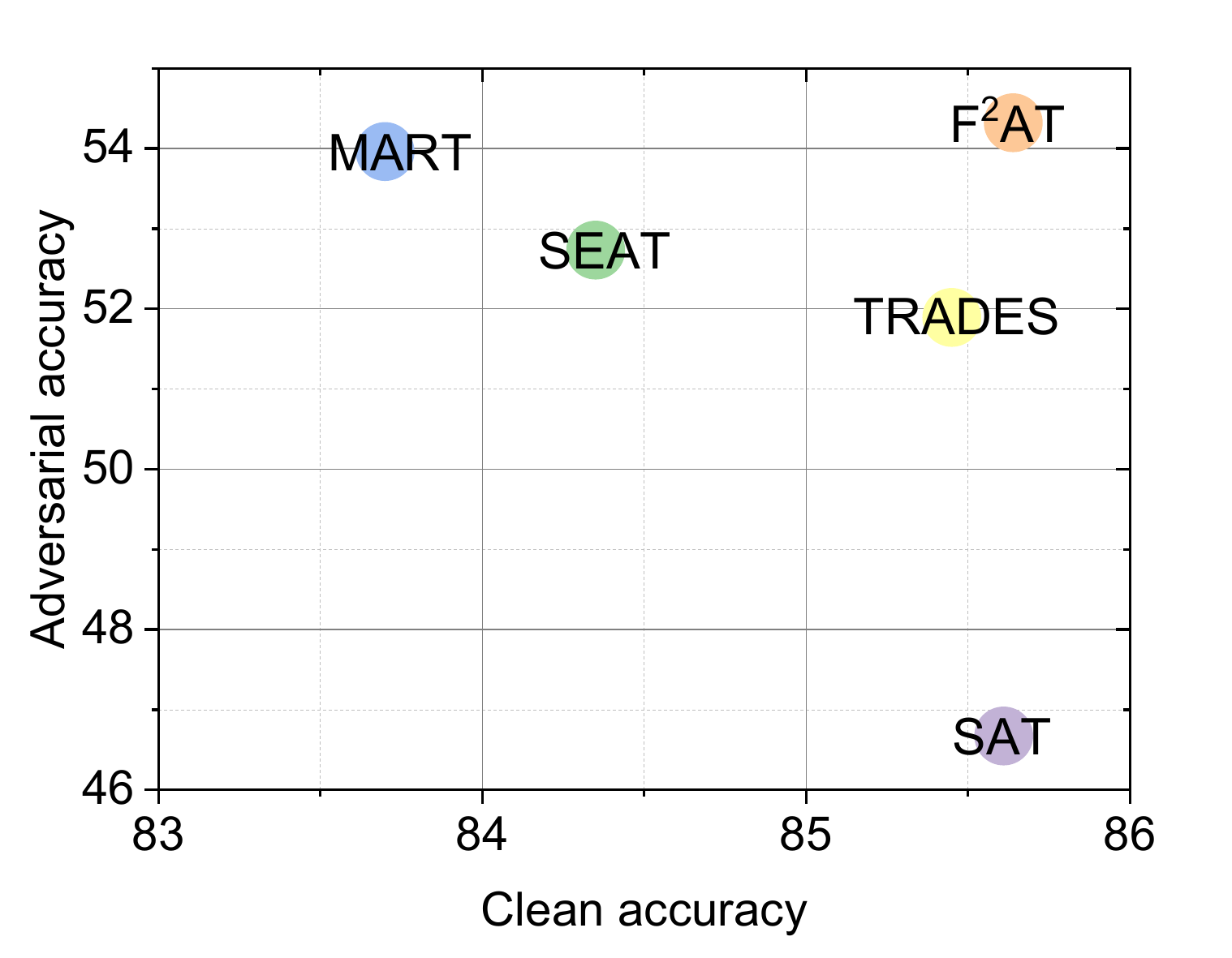}
		\caption{WRN-34-10}
		\label{WRN}
	\end{subfigure}
        \caption{Scatterplot of different methods regarding clean and robust accuracy. The robust accuracy is tested on the CIFAR-10 dataset generated by PGD20. The figure shows that our method achieves a better trade-off.}
        \label{acc_scatter}
        \end{figure}
        
	\textbf{Defense setting.}
	We configured the training process with 100 epochs, incorporating a transition epoch set at [45, 75]. This decision was based on empirical observations indicating that model performance stabilized before reaching the 100-epoch mark. During training, all images underwent standard data augmentation, including random crops with a 4-pixel padding and random horizontal flips \cite{Krizhevsky2009LearningML}. To ensure a fair comparison, we employed SGD for all methods, featuring a momentum of 0.9, a weight decay of $2\times10^{-4}$, and a batch size of 128. The initial learning rate is set to 0.1, reduced by a factor of 10 at the 50th and 75th epochs. Furthermore, we fine-tuned hyperparameter settings for the defense methods, establishing $K = 2$, $\gamma = 1$, $\tau = 0.07$, $\upsilon = 0.995$, and $\alpha = 0.1$. 

	\subsection{Performance under White-box Attacks}
	In the white-box setting, all attacks possess access to the architecture and parameters of the target model. We assess the robustness of CIFAR-10 and CIFAR-100 through various adversarial attacks, including PGD20, PGD100, MI-FGSM, CW, and AA. To ensure a thorough evaluation, we conducted multiple experiments using different random seeds, and the results are presented in Table \ref{tb1-CIFAR-10} to \ref{tb1-CIFAR-100-res}. These tables showcase the averaged outcomes obtained from these diverse experiments.
       
	 \textbf{CIFAR-10.} We perform experiments on CIFAR-10, which has 10 classes, 5K training images, and 10K test images per class. We report the accuracy of the original test images (Clean) and under 5 attacks. The findings, meticulously tabulated in Table \ref{tb1-CIFAR-10-res} and Table \ref{tb1-CIFAR-10}, showcase the efficacy of our approach in achieving robustness across diverse adversarial scenarios, all while maintaining a commendable clean accuracy of 80.56\% and 85.64\%. Fig. \ref{acc_scatter} shows that our method achieves a better trade-off. For the ResNet-18 architecture, our method demonstrates noteworthy resilience against the selected attacks, each exhibiting varying degrees of adversarial strength. Compared with SAT, the improvement in robustness accuracy for weak attacks (PGD20, PGD100, MI-FGSM) (5.20\% vs. 5.24\% vs. 3.73\%) is more significant than that for strong attacks (3.91\% vs. 3.61\%). This robustness improvement is even more pronounced in the WRN-34-10. 
  
    Compared with SAT, TRADES, MART, and SEAT, our method performs competitively in clean accuracy (85.64\% vs. 85.45\% vs. 83.70\% vs. 84.35\%). Notably, under the PGD20 attack, our method achieves an accuracy of 54.32\%, surpassing the closest competitor by 0.36\% and eclipsing SAT by an impressive 7.65\%. Moreover, our method consistently delivers superior gains in adversarial accuracy under PGD100, TI-FGSM, CW, and AA. We further explored the effects of the perturbation budget and the number of iteration steps (see Fig. \ref{epi-steps}). The results show that our method is more effective for small perturbations, and shows better results at any number of steps. Moreover, the prediction frequency of our method is more balanced for each class (See Fig. \ref{res-Frequency}). 
        \begin{table*}[t!]

		\centering
		\caption{Adversarial accuracy (\%) of defense methods against black-box attacks. The target model is WRN-34-10 and the surrogate model is adversarially trained VGG-16. The best results are
indicated in bold.}
  \resizebox{0.8\textwidth}{!}{
		\begin{tabular}{llccccccc} 
  
			\toprule
			Dataset& Defense & FGSM & PGD20 & TI-FGSM & MI-FGSM & DI-FGSM & VMI-FGSM\\
			\midrule
			 \multirow{5}{*}{CIFAR-10} &SAT \cite{DBLP:conf/iclr/MadryMSTV18} & 73.29 & 72.75 & 77.47 & 72.64 & 73.07 & 72.47 \\
			\multirow{5}{*}{} & TRADES \cite{DBLP:conf/icml/ZhangYJXGJ19} & 74.06 & 74.01 & 78.28 & 74.85 & \textbf{75.11} & 74.81 \\
            \multirow{5}{*}{} & MART\cite{DBLP:conf/iclr/0001ZY0MG20} & 71.45 & 71.19 & 75.62 & 70.90 & 71.31 & 70.71 \\
           
            \multirow{5}{*}{} & SEAT \cite{wang2022self} & 69.55 & 69.17 & 74.57 & 68.95 & 69.38 & 68.77 \\
            
			\multirow{5}{*}{} & F$^2$AT(ours) & \textbf{74.39} & \textbf{74.28} & \textbf{79.02} & \textbf{74.99} & 74.91 & \textbf{74.86} \\
            \midrule
			 \multirow{5}{*}{CIFAR-100} &SAT \cite{DBLP:conf/iclr/MadryMSTV18} &  38.70 & 38.57 & 41.69 & 38.38 & 38.64 & 38.24 \\
			
                \multirow{5}{*}{} & TRADES \cite{DBLP:conf/icml/ZhangYJXGJ19} & 46.39 & 46.31 & 48.48 & 46.10 & 46.21 & 46.07 \\
            \multirow{5}{*}{} & MART\cite{DBLP:conf/iclr/0001ZY0MG20} & 46.03 & 46.13 & 48.24 & 45.64 & 45.84 & 45.66 \\
            
            \multirow{5}{*}{} & SEAT \cite{wang2022self} & 43.82 & 43.50 & 46.82 & 43.42 & 43.53 & 43.28 \\
            
			\multirow{5}{*}{} & F$^2$AT(ours) & \textbf{46.62} & \textbf{46.81} & \textbf{49.81} & \textbf{46.65} & \textbf{46.71} & \textbf{46.33} \\
			\bottomrule
		\end{tabular}
  }
		\label{tb2-CIFAR-bb}
	\end{table*}

        \begin{table*}[t!]
        
		\centering
		\caption{Adversarial accuracy (\%) of defense methods against black-box attacks. The target model is ResNet-18 and the surrogate model is adversarially trained VGG-16. The best results are indicated in bold.}
  \resizebox{0.8\textwidth}{!}{
		\begin{tabular}{llccccccc} 
			\toprule
			Dataset& Defense & FGSM & PGD20 & TI-FGSM & MI-FGSM & DI-FGSM & VMI-FGSM\\
			\midrule
			 \multirow{5}{*}{CIFAR-10} &SAT \cite{DBLP:conf/iclr/MadryMSTV18}  & 61.72 & 61.16 & 65.92 & 60.89 & 61.27 & 60.48 \\
			\multirow{5}{*}{} & TRADES \cite{DBLP:conf/icml/ZhangYJXGJ19}  & 64.96 & 64.48 & 68.68 & 64.29 & 64.55 & 64.07 \\
            \multirow{5}{*}{} & MART\cite{DBLP:conf/iclr/0001ZY0MG20} & 63.50 & 63.00 & 67.93 & 62.78 & 63.19 & 62.61 \\
           
            \multirow{5}{*}{} & SEAT \cite{wang2022self} & 63.75 & 63.17 & 68.90 & 63.09 & 63.49 & 62.90 \\
            
			\multirow{5}{*}{} & F$^2$AT(ours) & \textbf{67.39} & \textbf{66.97} & \textbf{72.44} & \textbf{66.95} & \textbf{67.21} & \textbf{66.78} \\
            \midrule
			 \multirow{5}{*}{CIFAR-100} &SAT \cite{DBLP:conf/iclr/MadryMSTV18} & 37.40 & 37.15 & 40.23 & 37.02 & 37.26 & 36.84 \\
			
                \multirow{5}{*}{} & TRADES \cite{DBLP:conf/icml/ZhangYJXGJ19} & 40.07 & 39.82 & 42.52 & 39.08 & 39.78 & 39.45 \\
            \multirow{5}{*}{} & MART\cite{DBLP:conf/iclr/0001ZY0MG20} & 38.53 & 38.39 & 40.51 & 38.02 & 38.13 & 37.92 \\
            
            \multirow{5}{*}{} & SEAT \cite{wang2022self} & 38.80 & 38.39 & 41.90 & 38.22 & 38.35 & 37.97 \\
            
			\multirow{5}{*}{} & F$^2$AT(ours) & \textbf{40.20} & \textbf{39.99} & \textbf{43.54} & \textbf{39.66} & \textbf{39.88} & \textbf{39.60} \\
			
			\midrule
			 \multirow{5}{*}{Tiny-ImageNet} &SAT \cite{DBLP:conf/iclr/MadryMSTV18}  & 30.77 & 30.78 & 30.63 & 30.75 & 30.81 & 30.65 \\
			\multirow{5}{*}{} & TRADES \cite{DBLP:conf/icml/ZhangYJXGJ19}  & 35.78 & 35.81 & 35.75 & 35.78 & 35.79 & 35.59 \\
            \multirow{5}{*}{} & MART\cite{DBLP:conf/iclr/0001ZY0MG20} & 33.33 & 33.39 & 33.38 & 33.35 & 33.39 & 33.16 \\
           
            \multirow{5}{*}{} & SEAT \cite{wang2022self} & 35.45 & 35.51 & 35.55 & 35.44 & 35.54 & 35.45 \\
            
			\multirow{5}{*}{} & F$^2$AT(ours) & \textbf{40.55} & \textbf{40.64} & \textbf{40.51} & \textbf{40.43} & \textbf{40.58} & \textbf{40.46} \\
            
            \bottomrule
		\end{tabular}
  }
		\label{tb2-CIFAR-bb-res}
	\end{table*}
	\textbf{CIFAR-100.} We also conducted experiments on the CIFAR-100 dataset, with 100 classes, 50K training, and 10K test images. Note that this dataset is more challenging than CIFAR-10 because the number of training images per class is one-tenth of that of CIFAR-10. The results in Table \ref{tb1-CIFAR-100} and \ref{tb1-CIFAR-100-res} show that F$^2$AT performs better than other benchmarking methods. When using the ResNet-18 model architecture, our method demonstrates a noticeable improvement in recognition accuracy for AA attacks compared to other methods. Specifically, compared to SAT, our method achieves a 3.61\% increase in accuracy while maintaining clean accuracy levels. Furthermore, for WideResNet-34-10, the robustness improvement offered by our method is even more significant. In the case of PGD20, our method improves accuracy by 0.36\% compared to the second-highest method, while maintaining clean accuracy. Similarly, for AA attacks, our method demonstrates an improvement of 0.37\%. These results highlight our approach's effectiveness in enhancing the models' resilience against adversarial attacks.

        \textbf{Tiny-ImageNet} We further generalize our experiments to the Tiny-ImageNet dataset. This dataset has 200 classes. Each class has 500 training images, and 50 test images. As shown in
        Table. \ref{tb1-tiny-res}, Our method demonstrates enhanced robust accuracy while maintaining clean accuracy. Our approach shows a remarkable increase in clean and adversarial accuracy. Specifically, our clean accuracy has increased by 4.35\% compared to the highly accurate TRADES method. Furthermore, our method has achieved a 1.76\% improvement in robust accuracy compared to the second-highest performing approach for AA.
\begin{figure}[t]
		\centering
		\includegraphics[width=0.35\textwidth]{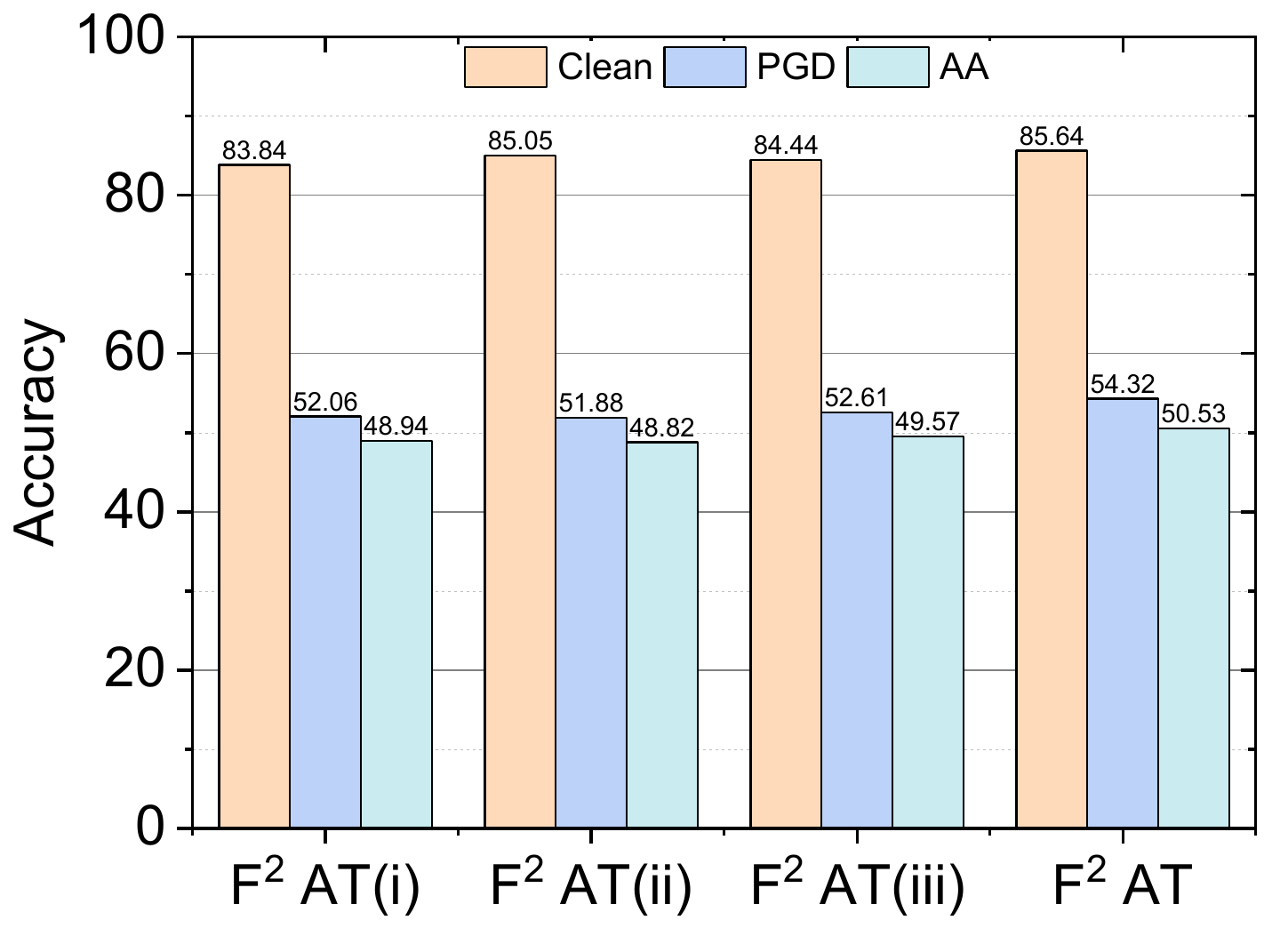}
		\caption{The ablation study on model WRN-34-10. F$^2$AT(i), (ii), (iii) corresponds to the following three scenarios in turns: (i) removing natural and perturbed patterns,  (ii) setting the hyperparameter $\gamma$ to 0 (Eq. \eqref{Loss_all}), (iii) setting the hyperparameter $\alpha$ to 0 (Eq. \eqref{Loss_all}).}
		\label{ablation}
	\end{figure}

 \begin{figure*}[t!]
		\centering
		\includegraphics[width=0.8\textwidth]{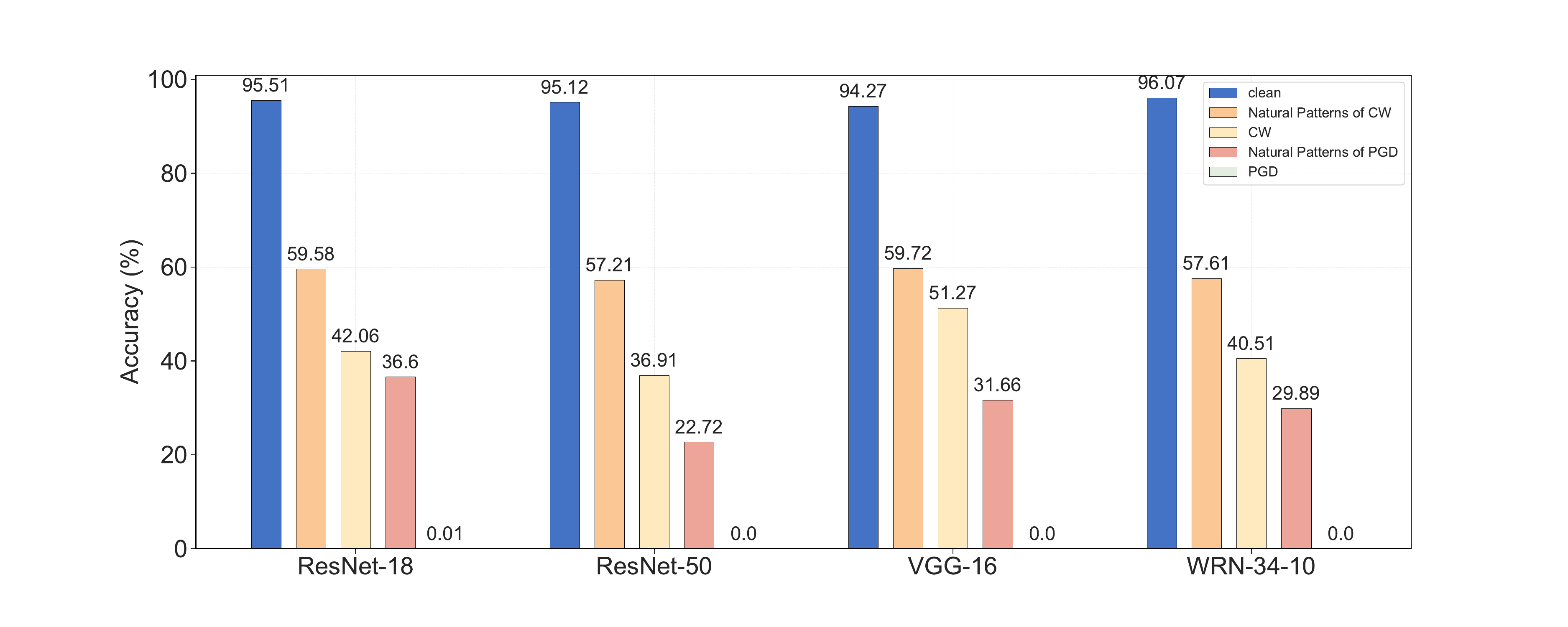}
		\caption{Histogram of the accuracy of different models (clean training) on natural patterns. We find that the model well-trained on clean examples has very low robustness to PGD attacks, with its highest robustness accuracy of 0.01\%, and slightly higher robustness to CW attacks, with a maximum of 51.27\%. In contrast, the accuracy of these models against the natural patterns of the adversarial examples is substantially improved, most notably by ResNet-18.}
		\label{ablation_pattern}
	\end{figure*}

   \begin{figure}[t]
		\centering
			\includegraphics[width=0.75\linewidth]{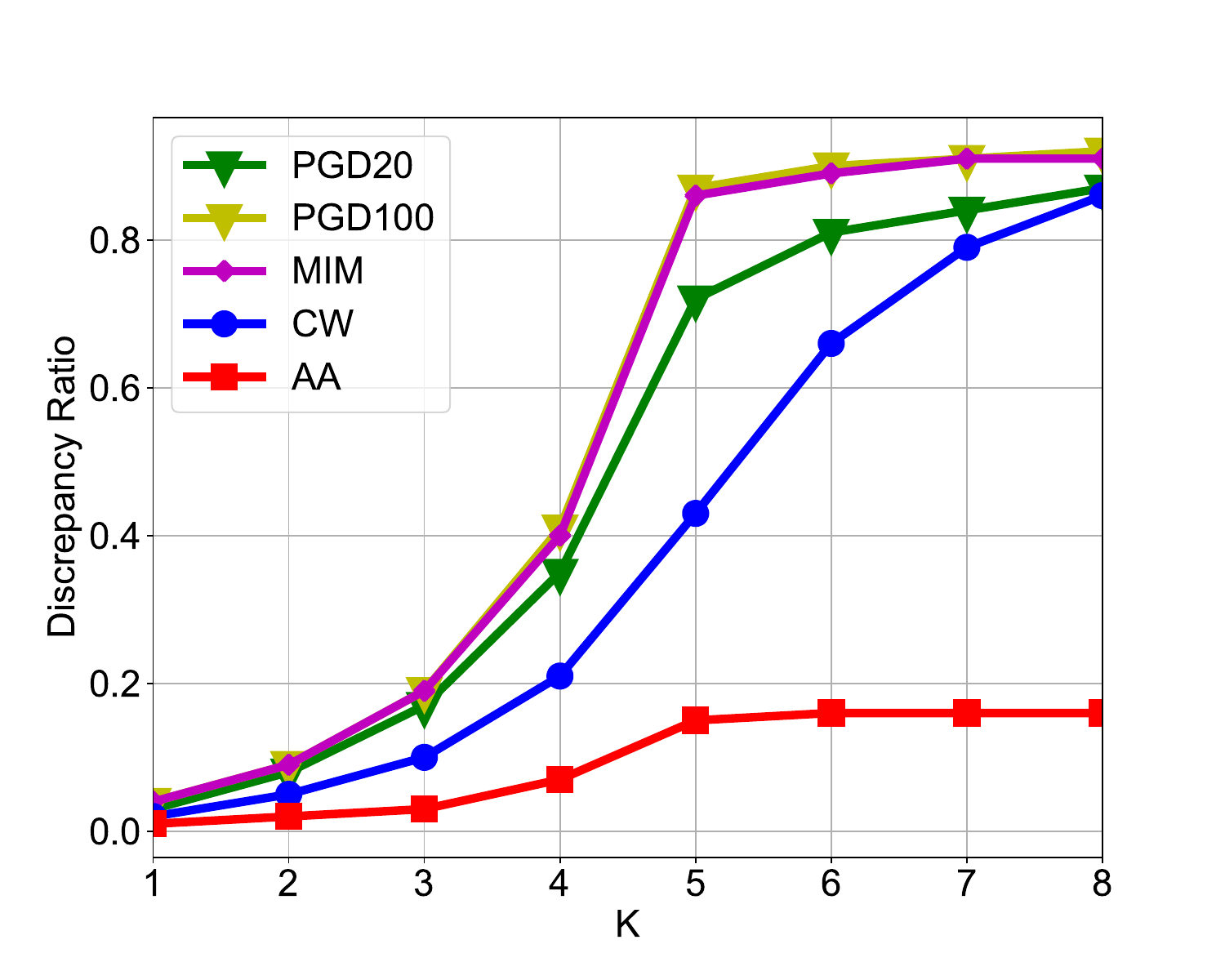}
			\caption{Discrepancy ratio of the natural patterns of adversarial examples and the natural patterns of the clean examples. The smaller the value of $K$, the smaller the difference between the natural patterns of the adversarial examples and the clean examples.}
			\label{discrepancy}
	\end{figure}

	\subsection{Performance under Black-box Attacks}
	Under a black-box attack, the attacker does not know anything about the internal structure of the attacked model, training parameters, defense methods, etc., and can only interact with the model through the output. The results presented in Table \ref{tb2-CIFAR-bb} and \ref{tb2-CIFAR-bb-res} demonstrate the superior robustness of our method compared to other baseline techniques. Specifically, when evaluating defense methods against black-box attacks, all the approaches exhibit significantly enhanced robustness compared to white-box scenarios. In the worst-case scenario across all attacks, our method F$^2$AT achieves robust accuracy gains of 6.30\%, 2.71\%, 4.17\%, and 3.88\% over SAT, TRADES, MART, and SEAT, respectively. These results highlight the effectiveness of our approach in mitigating adversarial attacks and elevating the resilience of the models in the black-box setting.

   \begin{figure*}[t!]
		\centering
            \begin{subfigure}{\linewidth}
                \centering
                \includegraphics[width=0.95\textwidth]{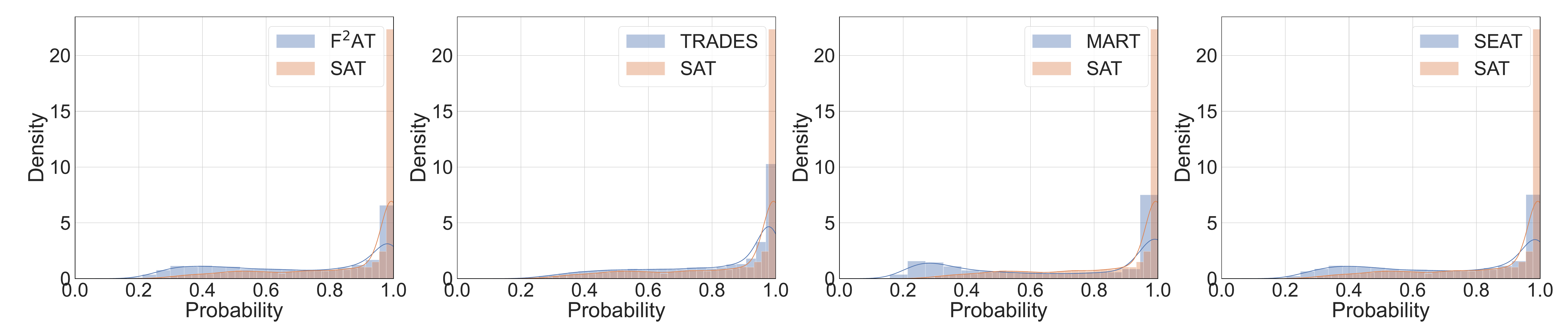}
                \caption{Clean}
            \end{subfigure}
            \centering
            \begin{subfigure}{\linewidth}
                \centering
                \includegraphics[width=0.95\textwidth]{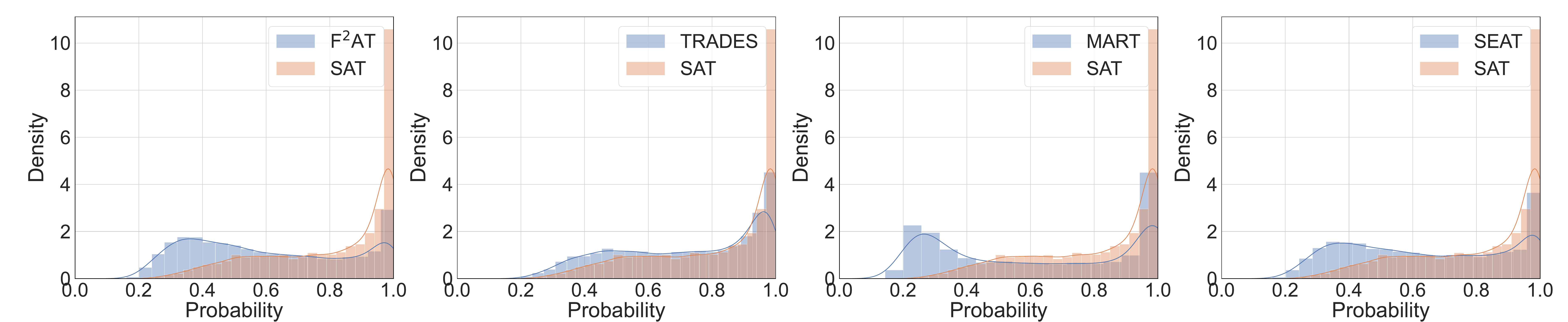}
                \caption{Adversarial}
            \end{subfigure}
		
	\caption{Plot of the predicted probability distribution for clean examples and adversarial examples. All adversarial examples are from the CIFAR-10 test set under a standard PGD attack. The color curve is the probability density of each method, which indicates that the prediction confidence distribution of our model is more stable compared to the others.}
		\label{hist}
	\end{figure*}

	\subsection{Ablation Study}
        To investigate the role of individual algorithmic components in improving adversarial robustness, we conducted the ablation study in three different cases: F$^2$AT(i) removing natural and perturbed patterns, F$^2$AT(ii) setting the hyperparameter $\gamma$ to 0 (Eq. \eqref{Loss_all}), F$^2$AT(iii) setting the hyperparameter $\alpha$ to 0 (Eq. \eqref{Loss_all}). We evaluated the performance of these variants using the PGD20 attack in CIFAR-10. As shown in Fig. \ref{ablation}, the results show that each component of our method contributes.


	\textbf{Impact of natural and perturbed patterns.}
    We begin by examining the significance of natural patterns. We conducted experiments using well-trained models with four different architectures (ResNet-18, ResNet-50, VGG-16, WRN-34-10) to classify adversarial images generated by PGD20 and CW on the CIFAR-10 dataset. The results are illustrated in Fig. \ref{ablation_pattern}. Upon scrutinizing the findings depicted in Fig. \ref{ablation_pattern}, our inference is that natural patterns play a crucial role in augmenting robustness. The removal of both natural and perturbed patterns leads to a loss of meaning in the loss function $\mathcal{L}_{pd}$. In such instances, we opt to set $\alpha$ to 0 and substitute clean examples instead of natural patterns. This substitution is deemed justifiable based on the observation that, when the value of $K$ is small, the discrepancy between the natural pattern of the adversarial example and that of the clean example is minimal (see Fig. \ref{discrepancy}). 
    \begin{table}[!t]
		\centering
		\caption{The effect of $K$ on the clean and adversarial accuracy of the model. The target model is ResNet-18, and the dataset used is CIFAR10. The smaller $K$ is, the higher the clean accuracy is, and the larger $K$ is, there is a small increase in adversarial accuracy.}
       
	\begin{tabular}{ccccccc}
        \toprule
         & Clean & PGD20 & PGD100 & MIM & CW & AA \\
        \midrule
        $K=2$ & \textbf{82.40} & 49.77 & 49.78 & 51.05 & 48.59 & 46.18 \\
        $K=4$ & 81.66 & 49.71 & 49.67 & 50.57 & 47.86 & 45.73 \\
        $K=6$ & 81.81 & 49.59 & 49.52 & 50.41 & 48.07 & 45.92 \\
        $K=8$ & 80.73 & \textbf{50.34} & \textbf{50.31} & \textbf{51.36} & \textbf{48.77} & \textbf{46.66} \\
        \bottomrule
        \end{tabular}
		\label{K-influence}
	\end{table}
    Comparatively, the clean accuracy and adversarial accuracy of the model trained under this condition experience varying degrees of degradation. Specifically, the clean accuracy decreases by 1.80\%, while the adversarial accuracy against PGD decreases by 2.26\%. This discrepancy in performance can be attributed to the presence of natural noise in clean images relative to natural patterns. Such noise exerts an influence on the model’s overall performance. 
    
    We further investigated the effect of natural and perturbed patterns determined by different values of $K$ on the accuracy of the model (See Table. \ref{K-influence}). We have observed that the clean accuracy of the model improves with smaller values of $K$, whereas larger values of $K$ lead to a decrease in clean accuracy but a slight improvement in robustness. Specifically, when $K=8$, the natural pattern of the adversarial sample is equivalent to itself. Comparing the scenarios of $K=2$ and $K=8$, we find that the inclusion of natural patterns can contribute to enhanced clean accuracy while maintaining the model’s robustness. These findings underscore the importance of natural and perturbed patterns in fostering improved model accuracy and robustness.
        
        \textbf{Impact of $\mathcal{L}_{mg}$.}
        The improvement of model clean accuracy and robust accuracy is mainly attributed to $L_{mg}$. When $\alpha=0$, $\gamma=0$, it is equivalent to SAT, the larger $\gamma$ is, the more the model tends to focus on the margins of natural patterns. When we set $\gamma= 0 $, the model may fail because of the lack of guarantees to make the model focus on discriminative class features. At this point, the clean accuracy decreased by 1.20\%, and the robust accuracy for PGD and AA decreased by 1.71\% and 0.96\%, respectively. To further explore the effect of Loss on clean and robust accuracy. We plotted the predicted probability distributions of clean and adversarial examples (see Fig. \ref{hist}). It is observed that the probability distributions of F$^2$AT are more stable than those of other methods. In other words, F$^2$AT does not classify adversarial examples with misleading perturbations to the ground truth label with a high probability, nor does it classify clean examples, which may be misleading, to the correct label with a high probability.
        
        \textbf{Impact of $\mathcal{L}_{pd}$.} We conducted further experiments by setting $\alpha=0$ and $\alpha=1$ to investigate the impact on the experimental results of $\mathcal{L}_{pd}$. Our findings revealed that when $\alpha=1$, during the epochs of 60 and 100, the values of $\mathcal{L}_{mg}$ and $\mathcal{L}_{ce}$ gradually approached 0, indicating that these losses were significantly smaller than $\mathcal{L}_{pd}$. It became evident that when the model already had a certain level of accuracy, setting $\alpha=1$ would cause the model to overly focus on the differences between natural and perturbed patterns, rather than prioritizing natural or adversarial accuracy. This could lead to issues such as the model failing to converge or achieving poor overall accuracy. Compared to F$^2$AT, our experiments (See In Fig. \ref{ablation}) showed that the robust accuracy of F$^2$AT(iii) for PGD and AA decreased by 1.71\% and 0.99\%, respectively. Additionally, the clean accuracy decreased by 1.20\%. This result is in line with our intuition of designing the loss: the model's greater attention to the differences between natural and disturbed modes helps to improve the model's clean and robust accuracies.

        Considering that the examples in the real scenario are not all adversarial, we retrain the model with our method on the clean examples and find that a clean accuracy of 93.83\% can still be achieved. This helps to generalize our method to datasets with a mixture of clean and adversarial examples.

	\section{Conclusion}
 
	In this paper, we propose a novel disentanglement way for adversarial examples by bit-plane slicing. In this way, the adversarial example can be disentangled into natural and perturbed patterns, which provides the possibility for adversarial training to focus on core features to improve clean accuracy and robustness. This differs from previous adversarial defense methods in that the regular term of F$^2$AT does not depend on clean examples but on the natural patterns of images. The experimental results showed that F$^2$AT can significantly improve the adversarial robustness and clean accuracy of the model. In future work, we will further explore the relationship between model robustness and natural perturbations of images. We hope an interpretable and high-accuracy trade-off to be found.


 
%

\bibliographystyle{IEEEtran}
\bibliography{F2AT}


 




\vfill

\end{document}